\documentclass[journal]{new-aiaa}
\usepackage[utf8]{inputenc}
\usepackage{textcomp}

\usepackage{graphicx}
\usepackage{epsfig}
\usepackage{amsmath}
\usepackage[version=4]{mhchem}
\usepackage{longtable,tabularx}
\usepackage{multirow}
\usepackage{colortbl}
\usepackage{bm}
\usepackage{color}
\usepackage{mathrsfs}
\usepackage{enumitem}

\renewcommand\arraystretch{0.8}

\title{Lifting-wing Quadcopter Modeling and Unified Control}

\author{Quan Quan\footnote{Corresponding Author, is with the School of Automation Science and Electrical Engineering, Beihang University, Beijing 100191, China. Email: qq\_buaa@buaa.edu.cn.}, Shuai Wang, and Wenhan Gao}

\begin{document}
	\begin{sloppypar}
\maketitle
\begin{abstract}
Hybrid unmanned aerial vehicles (UAVs) integrate the efficient forward flight of fixed-wing and vertical takeoff and
landing (VTOL) capabilities of multicopter UAVs. This paper presents the modeling, control and simulation of a new type of hybrid  micro-small UAVs, coined as lifting-wing quadcopters. The airframe orientation of the lifting wing needs to tilt a specific angle often within $ 45$ degrees, neither nearly $ 90$ nor approximately $ 0$ degrees. Compared with some convertiplane and tail-sitter UAVs, the lifting-wing quadcopter has a highly reliable structure, robust wind resistance, low cruise speed and reliable transition flight, making it potential to work fully-autonomous outdoor or some confined airspace indoor. In the modeling part, forces and moments generated by both lifting wing and rotors are considered. Based on the established model, a unified controller for the full flight phase is designed. The controller has the capability of uniformly treating the hovering and 
forward flight, and enables a continuous transition between two modes, depending on the velocity command. What is more, by taking rotor thrust and aerodynamic force under consideration simultaneously, a control allocation based 
on optimization is utilized to realize cooperative control for energy saving. Finally, comprehensive Hardware-In-the-Loop (HIL) simulations are performed to verify the advantages of the designed aircraft and the proposed controller.
\end{abstract}

\section*{Nomenclature}


{\renewcommand\arraystretch{1.0}
\noindent\begin{longtable*}{@{}l @{\quad=\quad} l@{}}
$o_{\rm e}x_{\rm e}y_{\rm e}z_{\rm e}$ & Earth-Fixed Coordinate Frame(${}^{\rm e}\mathcal{F} $) \\
$o_{\rm b}x_{\rm b}y_{\rm b}z_{\rm b}$ & Quadcopter-Body Coordinate Frame(${}^{\rm b}\mathcal{F} $) \\
$o_{\rm l}x_{\rm l}y_{\rm l}z_{\rm l}$ & Lifting-Wing Coordinate Frame(${}^{\rm l}\mathcal{F} $) \\
$o_{\rm w}x_{\rm w}y_{\rm w}z_{\rm w}$ & Wind Coordinate Frame(${}^{\rm w}\mathcal{F} $) \\
$\mathbf{^{\rm{e}}p}$ & Position in ${}^{\rm e}\mathcal{F} $ \\
$\mathbf{^{\rm{e}}v}$ & Velocity in ${}^{\rm e}\mathcal{F} $ \\
$\mathbf{^{\rm{b}}v_{\rm a}}$, $\mathbf{^{\rm{l}}v_{\rm a}}$ & Airspeed vector in ${}^{\rm b}\mathcal{F} $ and ${}^{\rm l}\mathcal{F} $, respectively \\
${^{\rm{e}}{\mathbf v}_{\rm w}}$ & Wind velocity in ${}^{\rm e}\mathcal{F} $ \\
$ V_{\rm a}$ & Airspeed \\
$\phi$,  $\theta$,  $\psi$ & Euler angles in ${}^{\rm b}\mathcal{F} $\\ 
 
$\omega_{x_{\rm{b}}}$, $\omega_{y_{\rm{b}}}$, $\omega_{z_{\rm{b}}}$ & Angular velocity in ${}^{\rm b}\mathcal{F} $\\
 
$\alpha $ & Angle of attack in ${}^{\rm l}\mathcal{F} $\\

$\beta $ & Sideslip angle in ${}^{\rm l}\mathcal{F} $\\

${{C}_{L}}$ & Aerodynamic lift coefficient \\

${{C}_{D}}$ & Aerodynamic drag coefficient \\
 
${{C}_{m}}$ & Aerodynamic pitch moment coefficient \\

${{C}_{Y}}$ & Aerodynamic lateral force coefficient \\
 
${{C}_{l}}$ & Aerodynamic roll moment coefficient \\
 
${{C}_{n}}$ & Aerodynamic yaw moment coefficient \\ 

$\kappa$ & Installation angle of the lifting wing \\

$\eta$ & Installation angle of the motor \\

$c$ & Mean chord of the lifting wing \\

$b$ & Wingspan of the lifting wing \\

$S$ & Area of the lifting wing \\
\end{longtable*}}

\section{Introduction}
\subsection{Why Lifting-wing Quadcopter}
Unmanned aerial vehicles (UAVs) have attracted lots of recent
attention due to their outstanding performances in many fields, such
as aerial photography, precision farming, and unmanned cargo. According
to \cite{SAEED201891}, UAV platforms are currently dominated by three types:
fixed-wing UAV, rotorcraft UAV, and their hybrid that integrates the
advantages of the first two. The hybrid UAVs have the capability of
Vertical Take-off and Landing (VTOL), which enables more accessible
grounding or holding by hovering. This might be mandated by authorities
in high traffic areas such as lower altitudes in the urban airspace.
Furthermore, hybrid UAVs are categorized into two types: convertiplane
and tail-sitter. A convertiplane maintains its airframe orientation
in all flight modes, but a tail-sitter is an aircraft that takes off
and lands vertically on its tail, and the entire airframe needs to
tilt nearly $ 90^\circ$ to accomplish forward flight \cite{SAEED201891,raj2019attitude}.

In March 2015, Google announced that the tail-sitter UAV for a packet
delivery service was scrapped, because it is still too difficult to
control in a reliable and robust manner according to the conclusion
came by the project leader \cite{wsj.com}. Some studies try to remedy
this by newly designed controllers \cite{ritz2018global}. However, unlike this way, we will study
a new type of hybrid UAV, coined as the \textit{lifting-wing quadcopter} \cite{xiao2021lifting,zhang2021performance}, to overcome the difficulty 
Google's Project Wing faced. A lifting-wing quadcopter is a quadcopter \cite{quan2017introduction,emran2018review} with a lifting wing installed at a 
specific mounting angle. During the flight, the quadcopter will provide thrust upward and forward simultaneously; and the lifting wing also 
contributes a lifting force partially.

\begin{figure}
	\centering
	\includegraphics[scale=1.0]{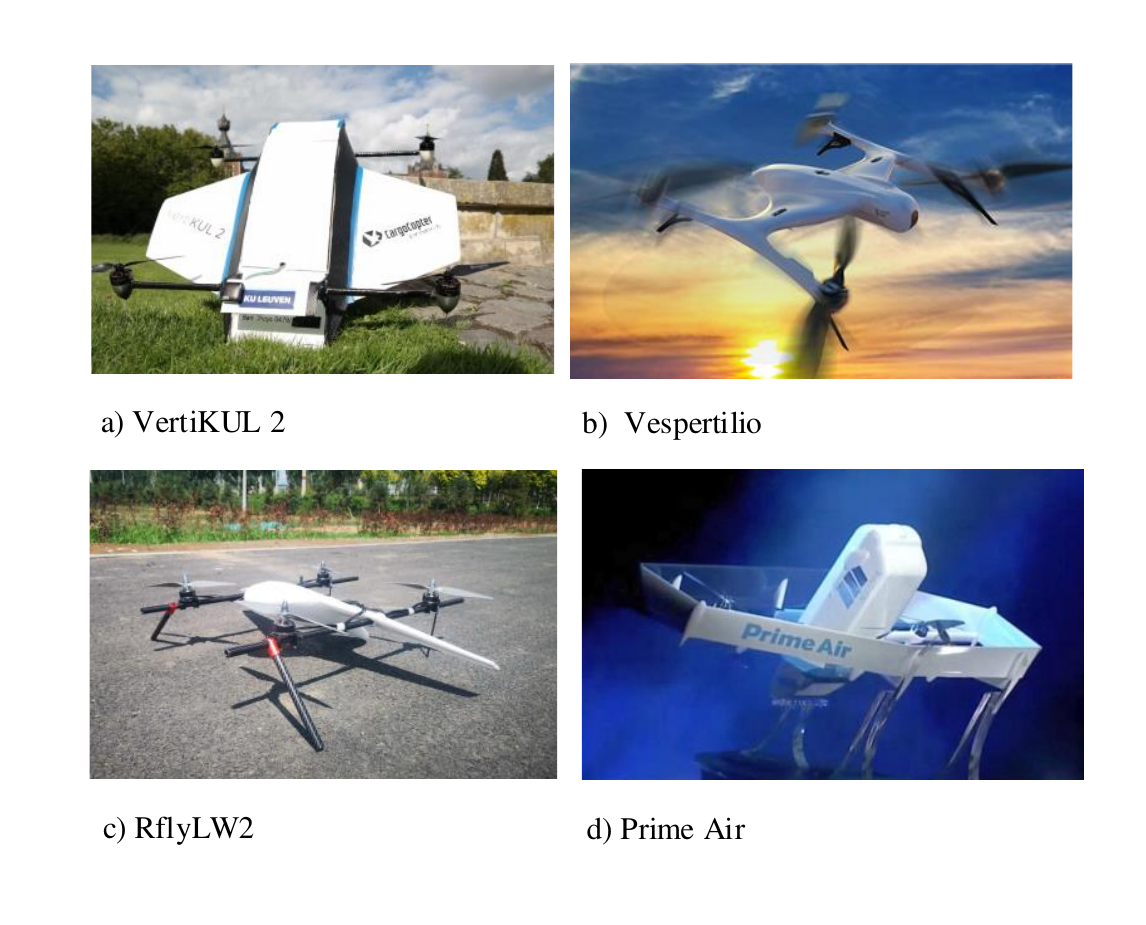}
	\caption{Prototypes of lifting-wing quadcopters.}
	\label{Fig:3}
\end{figure}

\begin{figure}
	\centering
	\includegraphics[scale=0.8]{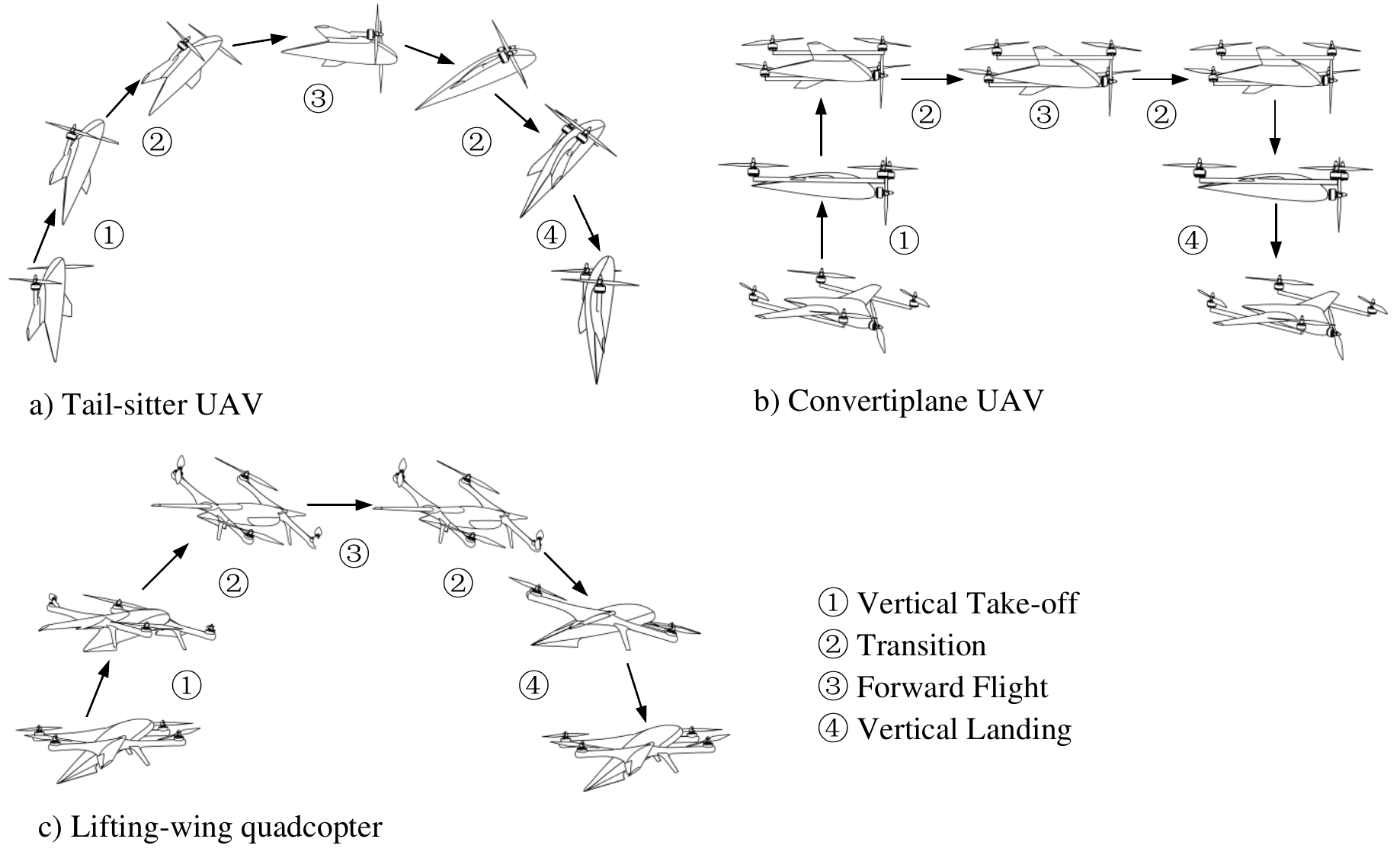}
	\caption{{Different flight modes of some VTOL UAVs.}}
	\label{Fig:2}
\end{figure}

As shown in Fig. \ref{Fig:3}, as far as we know, some prototypes of lifting-wing quadcopters in public can be found, such as VertiKUL2 by the University 
of Leuven (Fig. \ref{Fig:3} (a), Sept 2015)\cite{theys2016control}, Vespertilio by the VOLITATION company(Fig. \ref{Fig:3} (b))\cite{tx-tech.cn}, the 
latest version of the Prime Air delivery drone unveiled by Amazon(Fig. \ref{Fig:3}(d), Jun 2019)\cite{theverge.com} and, RflyLW2 by us(Fig. 
\ref{Fig:3} (c)) \cite{xiao2021lifting,zhang2021performance}.

The lifting-wing quadcopter is a new type of hybrid UAV because convertiplane and tail-sitter UAVs in their cruise phase work as fixed-wing UAVs, so the 
airframe orientation is defined as the head of the fixed-wing. But, the lifting-wing quadcopter is more like a quadcopter. The airframe orientation 
of the lifting wing needs to tilt a specific angle often within $ 45^\circ$, neither nearly $ 90^\circ$ (corresponding to tail-sitter UAVs) nor 
approximately $ 0^\circ$(corresponding to convertiplane UAVs). Fig. \ref{Fig:2} shows the full flight phase of the three VTOL UAVs. The design and 
performance evaluation of lifting-wing quadcopters have been studied extensively in \cite{xiao2021lifting,zhang2021performance}. In order to make this paper self-contained, we briefly introduce the advantages of the lifting-wing quadcopter compared with some convertiplane and tail-sitter UAVs.
\begin{itemize}
	\item \textbf{Highly reliable structure}. It does not require extra transition actuators.
	This is a reliable structure by eliminating the need for complicated
	control.
	\item \textbf{Robust wind resistance}. It has a shorter lifting wing compared with
	the corresponding fixed wing of convertiplanes and tail-sitter UAVs, because rotors can share the lift. Moreover, as
	shown in Fig. \ref{Fig:8}, it does not have a vertical rudder. Instead,
	this function is replaced by the yaw control of the quadcopter component.
	In order to improve the yaw control ability, the axes of rotors do
	not point only upward anymore as shown in Fig. \ref{Fig:8}(a). This implies
	that the thrust component by rotors can change the yaw directly rather
	than merely counting on the reaction torque of rotors. From the above,
	the wind interference is significantly reduced on the one hand; on
	the other hand, the yaw control ability is improved. As a result,
	it has better maneuverability and hover control to resist the
	disturbance of wind than those by tail-sitter and convertiplane UAVs.
	\item \textbf{Low cruise speed}. It can make a cruise at a lower speed than that by convertiplanes and tail-sitter UAVs, meanwhile
	saving energy compared with corresponding quadcopters. This is very
	useful when a UAV flies in confined airspace such as a tunnel, where the high speed
	is very dangerous. Although current hybrid UAVs can have a big or
	long wing for low cruise speed, they cannot work in many confined
	airspace due to their long wingspan. However, the lifting-wing quadcopter
	can be small.
	\item \textbf{Reliable transition flight}. When a tail-sitter UAV performs transition
	flight, the velocity and angle of attack will change dramatically,
	leading to complicated aerodynamics even stall. Unlike tail-sitter
	UAVs, the lifting-wing quadcopter only has to tilt a specific angle
	often smaller than $ 45^\circ$ rather than $ 90^\circ$. The airflow
	on the lifting wing is stable, and lift and drag are changed linearly
	with the angle of attack. These will avoid great difficulty (or say danger) in controlling within the full flight envelope.
\end{itemize}

\begin{figure}
	\centering
	\includegraphics[scale=0.5]{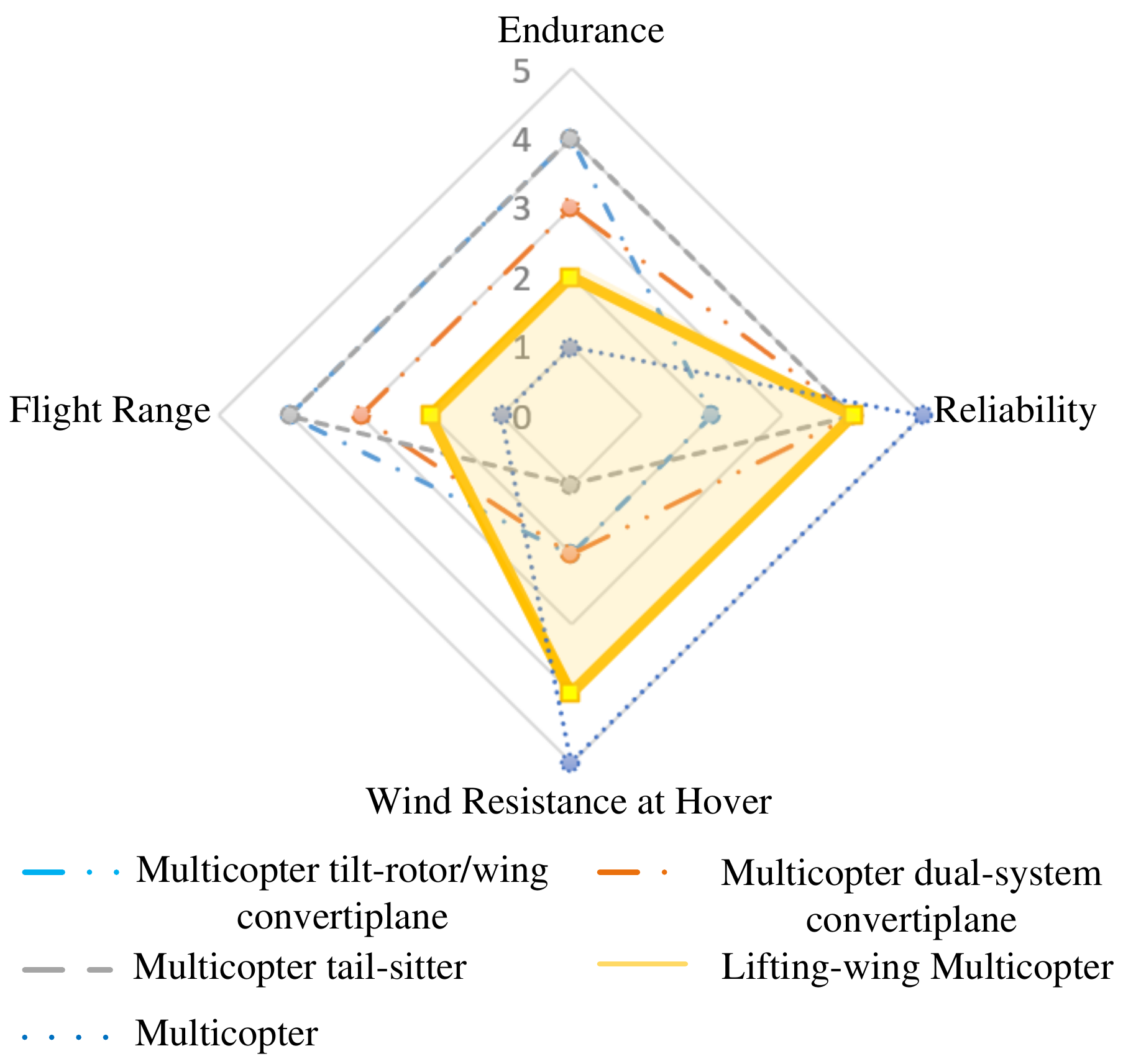}
	\caption{Comparison of different VTOL UAVs.}
	\label{Fig:4}
\end{figure}

With the four features above, it is potential for the lifting-wing quadcopter to work fully-autonomously outdoors or in some confined airspace indoors replacing with corresponding quadcopters.
The further comparisons with multicopter, tilt-rotor/wing convertiplane, multicopter dual-system convertiplane and multicopter tail-sitter \cite{raj2019attitude} are summarized in Tab. \ref{Tab:2} and Fig. \ref{Fig:4}.
As shown, the lifting-wing quadcopter possesses the feature between current hybrid UAVs and quadcopters, and it is more like a quadcopter.

\subsection{Control of Current Hybrid UAVs}
The control of the lifting-wing quadcopter has the following two distinguishing features.

\begin{table}
	\centering
	\caption{Comparison of different VTOL UAVs.}
	
	\begin{tabular}{|c|c|c|c|c|}
		\hline 
		& Endurance & Reliability & Wind Resistance at Hover & Flight Range\tabularnewline
		\hline 
		Multicopter tilt-rotor/wing convertiplane & 4 & 2 & 2 & 4\tabularnewline
		\hline 
		Multicopter dual-system convertiplane & 3 & 4 & 2 & 3\tabularnewline
		\hline 
		Multicopter tail-sitter & 4 & 4 & 1 & 4\tabularnewline
		\hline 
		\rowcolor{yellow}
		\hline
		Lifting-wing multicopter & 2 & 4 & 4 & 2\tabularnewline
		\hline 
		Multicopter & 1 & 5 & 5 & 1\tabularnewline
		\hline
		
	\end{tabular}
	\begin{flushleft}\qquad\qquad\  Note: bigger number implies better.\end{flushleft}
	\label{Tab:2}
\end{table}

\begin{itemize}
	\item \textbf{Unified control for full flight phases}. Hybrid UAVs often have three
	different flight modes, including the hover, the transition flight,
	and the forward flight. By taking the multicopter tilt-rotor/wing convertiplane,
	multicopter dual-system convertiplane and multicopter-tail sitter, for
	example, their take-off and landing are controlled only by the quadcopter
	component, while the forward flight is controlled like a fixed-wing
	aircraft. The two control ways are very different, so the transition
	flight is challenging due to the nonlinearities and uncertainties.
	However, a full flight phase of the lifting-wing quadcopter always involves thrust
	by the quadcopter and aerodynamic force by the lifting wing. Therefore,
	the lifting-wing quadcopter can be considered under only the transition
	flight mode in the full flight phase (hover control here also will
	take the aerodynamic force into consideration due to wind on the lifting wing). As a
	result, a unified control is needed. Fortunately, the lifting-wing
	quadcopter only needs to tilt a specific angle often smaller than
	$ 45^\circ$, rather than $ 90^\circ$ like tail sitter UAVs. This reduces
	the possibility of having a stall.
	\item \textbf{Cooperative control for energy saving}. The transition flight for current
	hybrid UAVs is very short, so not too much attention needs to pay to energy consumption in practice. However, it should be considered for the lifting-wing
	quadcopter as it is under the transition flight mode in the full
	flight phase. Cooperative control for energy saving is feasible. For
	example, roll control can be performed by both the quadcopter
	component and the ailerons by the lifting wing. Obviously, the aileron
	control is more energy-saving.
\end{itemize}

Among the control phase of a hybrid UAV, the transition control is the most challenging issue \cite{hassanalian2017classifications}, especially for tail-sitter UAVs. Since the
actuators of tail-sitter UAVs are like those of lifting-wing quadcopters,
the existing transition control of tail-sitter UAVs can be used for reference.

\begin{enumerate}[label=(\roman*)]
	\item \textbf{Trajectory open-loop control for transition flight}. 
	
	The open-loop trajectory tracking control is very straightforward. The principle is to make the UAV enter into another mode's condition by focusing on controlling
	some variables like altitude other than the trajectory, then switch
	to the controller of the next mode. For example, increasing
	thrust and reducing its pitch angle at the same time can make a tail-sitter
	UAV enter into forwarding flight \cite{argyle2013vertical,escareno2007modeling}. The aim is to keep the altitude the same 
	\cite{escareno2007modeling}. Because the transition time for tail-sitter UAVs is short, the trajectory will not change too much. Obviously, this 
	method is inapplicable to lifting-wing quadcopters.
	\item \textbf{Trajectory closed-loop control for transition flight}.
	
	\begin{itemize}
		\item \textit {Linearization method based on optimization}. According to the trajectory
		and the model, the reference state and feedforward are derived by
		optimization in advance \cite{oosedo2017optimal,li2020transition}. Based on them, the linearization can be performed. With the resulting 
		linear model, existing controllers
		against uncertainties and disturbance are designed \cite{li2018nonlinear,smeur2020incremental}.
		As for the lifting-wing quadcopter, this method is applicable when
		the model and transition trajectory are known prior. Furthermore,
		cooperative control of the quadcopter component or the ailerons of
		the lifting wing can be performed by taking energy-saving into optimization.
		However, in practice, the model is often uncertain as the payload
		is often changed, such as parcel delivery. Also, this method is not
		very flexible due to that the trajectory has to be known a priori.
		\item \textit {Nonlinear control method}. One way is to take all aerodynamic forces
		as disturbances, and only the quadcopter component works for the flight
		transition \cite{lyu2018simulation,liu2018robust}. This requires that the quadcopter has a
		strong control ability to reject the aerodynamic force. Another way
		takes the aerodynamic force into consideration explicitly to generate
		a proper attitude \cite{zhou2017unified,flores2018simple}. How to cooperatively control the
		quadcopter component and the actuators of fixed-wing is not found
		so far. This is because, we guess, the transition flight is often
		short, and more attention is paid to making the UAV stable by reducing
		the possibility of stall rather than optimization.
	\end{itemize}
\end{enumerate}

As shown above, the linearization method based on optimization is
somewhat not flexible, but cooperative control for energy saving can
be performed in an open-loop manner. The nonlinear control method
is flexible but not considering how to control cooperatively for energy
saving.

\subsection{Our Work and Contributions}
In this paper, we will consider designing a unified controller for the
full flight phase of a lifting-wing quadcopter. What is more, the
quadcopter component and the ailerons of the lifting wing work cooperatively
to save energy. First, we build the model of the lifting-wing quadcopter.
Unlike the tail-sitter UAV, it does not have a rudder, and its tilted
rotors will generate force components on the XY-plane in the quadcopter-body coordinate frame
(traditional quadcopters do not have the force component on the XY-plane).
Because of this, the translational dynamic involves
five control variables, namely three-dimensional force in the quadcopter-body coordinate frame
and two Euler angles (pitch and roll angles), further by considering the aerodynamic force determined
by Euler angles. However, it is difficult and a bit too early
to determine the five control variables according to the three-dimensional
desired acceleration, because it is hard to obtain the bounds of these
control variables. An improper choice may not be realized by actuators.
To this end, we only choose the $o_{\textrm{b}}z_{\textrm{b}}$ force
(the main force component) in the quadcopter-body coordinate frame and two Euler angles (pitch
and roll angles) to determine the desired acceleration uniquely, leaving
the other two force components as a lumped disturbance. This adopts the controlling
idea of quadcopters \cite{quan2017introduction,nascimento2019position}, but the computation method is
different due to the existence of the aerodynamic force. With the determined
Euler angles, moments are further determined in the lifting wing coordinate frame. So far, the unified control
for the full flight phase is accomplished. Finally, we will utilize the
control allocation to realize cooperative control for energy saving.
The $o_{\textrm{b}}z_{\textrm{b}}$ force and three-dimensional moments
will be realized by four rotors and two ailerons. This is why we have the
freedom to optimize the allocation for saving energy. The principle
behind this is to make the aerodynamic force (two ailerons) undertake the
control task as much as possible because aileron control is more energy-saving
than rotor control. As a result, cooperative control for energy saving
is accomplished.

The contributions of this paper are: (i) establish the model of a lifting-wing
quadcopter \textit{for the first time}; (ii) a unified controller design for the
full flight phase of the lifting-wing quadcopter; (iii) control allocation
for energy-saving performance. Comprehensive HIL simulation experiments are performed to show (i) the proposed lifting-wing quadcopter is more 
energy-saving with aileron; (ii) synthesizing the angular rate command from the coordinated turn in high-speed flight can reduce sideslip; (iii) the transition phase of the proposed lifting-wing quadcopter is significantly better than the tail-sitter and UAVs.

\section{Coordinate Frame}

\begin{figure}
	\centering
	\includegraphics[scale=0.95]{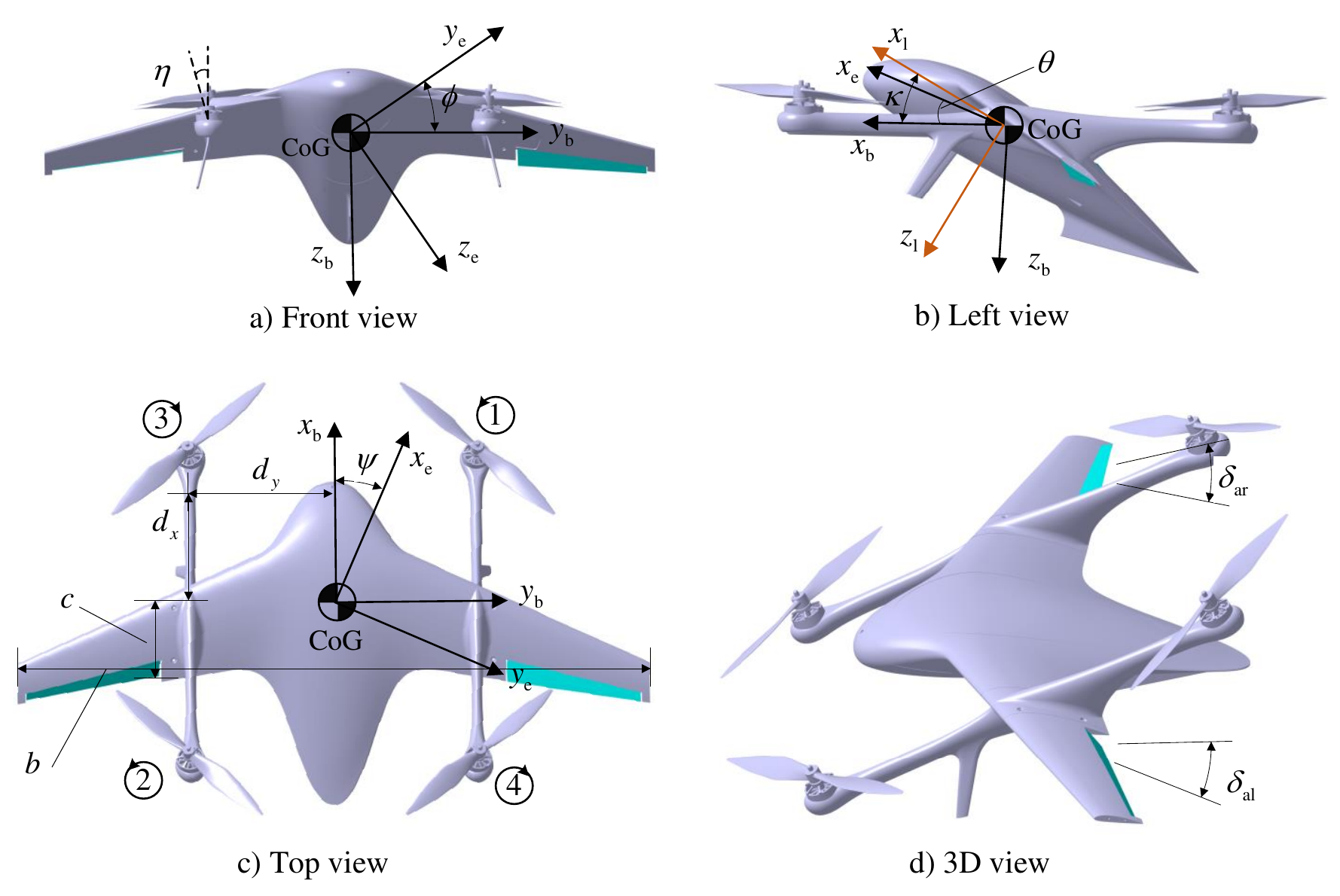}
	\caption{Coordinate frames and nomenclatures.}
	\label{Fig:8}
\end{figure}

A lifting-wing quadcopter is divided into two components, the lifting-wing component and the quadcopter component as shown in Fig. \ref{Fig:8}. 
According to these, the following coordinate frames are defined.

\subsection{Earth-Fixed Coordinate Frame (${}^{\rm e}\mathcal{F} $)}

The earth-fixed coordinate frame ${{o}_{\rm{e}}}{{x}_{\rm{e}}}{{y}_{\rm{e}}}{{z}_{\rm{e}}}$
is an inertial frame. The ${{o}_{\rm{e}}}{{z}_{\rm{e}}}$
axis points perpendicularly to the ground, and the ${{o}_{\rm{e}}}{{x}_{\rm{e}}}$
axis points to a certain direction in the horizontal plane. Then,
the ${{o}_{\rm{e}}}{{y}_{\rm{e}}}$ axis is determined according
to the right-hand rule. This frame is fixed, the initial position of the
lifting-wing quadcopter or the center of the Earth is often set as
the coordinate origin ${{o}_{\rm{e}}}$.

\subsection{Quadcopter-Body Coordinate Frame (${}^{\rm b}\mathcal{F} $)}

The quadcopter-body coordinate frame ${{o}_{\rm{b}}}{{x}_{\rm{b}}}{{y}_{\rm{b}}}{{z}_{\rm{b}}}$
is fixed to the quadcopter component of a lifting-wing quadcopter. The
Center of Gravity (CoG) of the lifting-wing quadcopter is chosen as the
origin ${{o}_{\rm{b}}}$ of ${}^{\rm b}\mathcal{F} $.
The ${{o}_{\rm{b}}}{{x}_{\rm{b}}}$ axis points to the nose
direction in the symmetric plane of the quadcopter. The ${{o}_{\rm{b}}{{z}_{\rm{b}}}
}$ axis is in the symmetric plane of the quadcopter, pointing downward,
perpendicular to the ${{o}_{\rm{b}}}{{x}_{\rm{b}}}$ axis.
The ${{o}_{\rm{b}}}{{y}_{\rm{b}}}$ axis is determined according
to the right-hand rule.

\subsection{Lifting-Wing Coordinate Frame (${}^{\rm l}\mathcal{F} $ )}

The lifting-wing coordinate frame ${{o}_{\rm{l}}}{{x}_{\rm{l}}}{{y}_{\rm{l}}}{{z}_{\rm{l}}}$
is fixed to the lifting-wing component. The origin ${{o}_{\rm{l}}}$ of ${}^{\rm l}\mathcal{F} $ is also set at the CoG of the lifting-wing quadcopter. The 
${{o}_{\rm{l}}}{{x}_{\rm{l}}}$ axis is in the symmetric plane pointing to the nose of the lifting wing. The ${{o}_{\rm{l}}}{{z}_{\rm{l}}}$ axis is in 
the symmetric plane of the lifting wing, pointing downward, perpendicular to the ${{o}_{\rm{l}}}{{x}_{\rm{l}}}$ axis, and the 
${{o}_{\rm{l}}}{{y}_{\rm{l}}}$ axis is determined according to the right-hand rule. The installation angle of the lifting wing, that is the angle 
between the $o_{\rm l}x_{\rm l} $ axis and the $ o_{\rm l}x_{\rm l}y_{\rm l}$ plane, is denoted by $\kappa\in{{\mathbb{R}}}$ as shown in Fig. 
\ref{Fig:8}(b).

\subsection{Wind Coordinate Frame (${}^{\rm w}\mathcal{F} $)}

The origin ${{o}_{\rm{w}}}$ of the wind coordinate frame ${{o}_{\rm{w}}}{{x}_{\rm{w}}}{{y}_{\rm{w}}}{{z}_{\rm{w}}}$
is also at the CoG of the lifting-wing quadcopter. The
${{o}_{\rm{w}}}{{x}_{\rm{w}}}$ axis is aligned with the airspeed vector. The ${{o}_{\rm{w}}}{{z}_{\rm{w}}}$
axis is in the symmetric plane of the lifting wing, pointing downward,
perpendicular to the ${{o}_{\rm{w}}}{{x}_{\rm{w}}}$ axis,
and the ${{o}_{\rm{w}}}{{y}_{\rm{w}}}$ axis is determined
according to the right-hand rule. The angle of attack (AoA), denoted by $\alpha \in{{\mathbb{R}}}$, is defined as the angle between the projection of the airspeed vector on the ${{o}_{\rm{l}}}{{x}_{\rm{l}}}{{z}_{\rm{l}}}$ plane and the ${{o}_{\rm{l}}}{{x}_{\rm{l}}}$ as shown in Fig. \ref{Fig:8}(b). The sideslip
angle, denoted by $\beta \in{{\mathbb{R}}}$, is defined as the angle between the airspeed vector and the ${{o}_{\rm{l}}}{{x}_{\rm{l}}}{{z}_{\rm{l}}}$ plane as shown in Fig. \ref{Fig:8}(c).

To convert the aerodynamic forces and moments acting on frame ${}^{\rm w}\mathcal{F} $ and ${}^{\rm l}\mathcal{F} $
to ${}^{\rm b}\mathcal{F} $ respectively, two rotation matrices are defined as followed:
\begin{eqnarray*}
	\mathbf{R}_{\text{w}}^{\text{b}}(\lambda )  = \left[\begin{matrix}\cos \lambda \cos \beta & -\cos \lambda \sin \beta & \sin\lambda\\
		\sin \beta & \cos\beta & 0\\
		{ - \sin \lambda \cos \beta } & \sin \lambda \sin \beta & {\cos \lambda }
	\end{matrix}\right], 
	{\mathbf{R}}_{\text{l}}^{\text{b}}  = \left[\begin{matrix}\cos\kappa & 0 & \sin\kappa\\
		0 & 1 & 0\\
		-\sin\kappa & 0 & \cos\kappa
	\end{matrix}\right],
\end{eqnarray*}
where $\lambda=\kappa-\alpha$. And the rotation matrix ${\mathbf{R}}_{\text{b}}^{\text{e}}$ maps a vector from frame ${}^{\rm b}\mathcal{F} $ to 
${}^{\rm 
e}\mathcal{F} $, 
defined by
\begin{equation*}
	\mathbf{R}_{\rm b}^{\mathrm{e}} =\left[\begin{array}{ccc}
		\cos \theta \cos \psi - \sin \theta \sin \phi \sin \psi& -\sin \psi \cos \phi & \cos \psi \sin \theta +\cos \theta \sin\phi \cos\psi\\
		\sin \theta \sin \phi \cos \psi+\cos \theta \sin \psi & \cos \phi \cos \psi & \sin \psi \sin \theta -\cos \phi\cos \theta \sin \phi \\
		-\cos \phi\sin \theta & \sin \phi& \cos \phi \cos \theta
	\end{array}\right].
\end{equation*}
\section{MODELING}

\subsection{Assumptions}

For the sake of model simplicity, the following assumptions are made:

\textbf{Assumption 1.} The body structure is rigid and symmetric about the $o_{\rm l}x_{\rm l}y_{\rm l} $ plane. 

\textbf{Assumption 2.} The mass and the moments of inertia are constant.

\textbf{Assumption 3.} The geometric center of the lifting-wing quadcopter
is the same as the CoG.

\textbf{Assumption 4. }The aircraft is only subjected to gravity,
aerodynamic forces, and the forces generated by rotors.

\subsection{Flight Control Rigid Model}
By \textbf{Assumptions 1-2}, the Newton's equation of motion is applied to get the translational motion as follows
\begin{equation}
	\begin{aligned}
		^{\rm{e}}\dot{{\mathbf p}} &={}^{\rm{e}}\mathbf{v} \\
		^{\rm{e}}\dot{{\mathbf v}} &=\mathbf{R}_{\rm{b}}^{\rm{e}}\frac{^{\rm{b}}\mathbf{f}}{m} \\
	\end{aligned}
	\label{Eq:position kinematic and dynamic}
\end{equation}
where $^{\rm{e}}\mathbf{p}=\left[{p_{x_{\rm e}}}\ \ {{p}_{{{y}_{\rm{e}}}}}\ \ {{p}_{{{z}_{\rm{e}}}}}\right]^{\text{T}}$ and 
${^{\rm{e}}\mathbf{v}}=\left[{{v}_{{{x}_{\rm{e}}}}}\ \ {{v}_{{{y}_{\rm{e}}}}}\ \ {{v}_{{{z}_{\rm{e}}}}}\right]^{\text{T}}$ are the position 
and velocity expressed in frame ${}^{\rm e}\mathcal{F} $ respectively; $m$ is the mass, $ {}^{\rm b}{\mathbf f} $ is the total force acting on the 
airframe expressed in frame ${}^{\rm b}\mathcal{F} $. 

To facilitate attitude control and combine the control characteristics of the rotor and lifting wing, the rotational dynamics is carried out in frame 
${}^{\rm l}\mathcal{F}$. It is given by Euler's equation of motion as 
\begin{equation}
	\begin{aligned}
		{\dot{\mathbf R}}_{\text{l}}^{\text{e}}&={\mathbf R}_{\text{l}}^{\text{e}}{\left[ {{}^{\text{l}}{\bm{\omega }}} \right]_ \times } \\
		{\mathbf{J}}\cdot{}^{\rm{l}}{\dot{\bm \omega}} &={}^{\rm{l}}\mathbf{m}-{}^{\rm{l}}{\bm \omega}\times(\mathbf{J\cdot}{}^{\rm{l}}{\bm \omega})
	\end{aligned}
	\label{Eq:attitude kinematic and dynamic}
\end{equation}
where the rotational matrix is derived by ${\mathbf R}_{\text{l}}^{\text{e}}={\mathbf R}_{\text{b }}^{\text{e}} {\mathbf R}_{\text{l}}^{\text{b 
}}$, ${\mathbf R}_{\text{l}}^{\text{b }} $ being a constant matrix; ${}^{\rm l}{\mathbf m} $ is the 
total moment acting on the airframe 
expressed in frame ${}^{\rm 
l}\mathcal{F} $, 
$^{\rm{l}}\bm{{\omega}}=\left[{{\omega}_{{{x}_{\rm{l}}}}}\ \ 
{{\omega}_{{{y}_{\rm{l}}}}}\ \ {{\omega}_{{{z}_{\rm{l}}}}}\right]^{\text{T}}$ is the angular velocity in frame ${}^{\rm l}\mathcal{F} $, ${\left[ 
{{}^{\text{l}}{\bm{\omega }}} \right]_ \times }$ denotes the skew-symmetric matric 
\begin{equation*}
	{\left[{{}^{\text{l}}{\bm{\omega }}} \right]_ \times }=\left[\begin{matrix}
		0 & -\omega _{z_{\rm l}} & \omega _{y_{\rm l}}\\
		\omega _{z_{\rm l}} & 0 & -\omega _{x_{\rm l}}\\
		-\omega _{y_{\rm l}} & \omega _{x_{\rm l}} & 0
	\end{matrix}\right],
\end{equation*}
and $\mathbf{J}\in{{\mathbb{R}}^{3\times3}}$ is the inertia matrix given by
\begin{equation*}
	\mathbf{J}=\left[\begin{matrix}{{J}_{x}} & 0 & -{{J}_{xz}}\\
		0 & {{J}_{y}} & 0\\
		-{{J}_{xz}} & 0 & {{J}_{z}}
	\end{matrix}\right].
\end{equation*}

\subsection{Forces and Moments}

By \textbf{Assumptions 3-4}, the total forces and moments acting on the UAV are decomposed into three parts: the 
aerodynamic forces and moments acting on the airframe (${{\mathbf{f}}_{\rm a}}$ and ${{\mathbf{m}}_{\rm a}}$),
the forces and moments generated by rotors (${{\mathbf{f}}_{\rm r}}$ and ${{\mathbf{m}}_{\rm r }}$), and the gravitational forces ${{\mathbf f_{g}}}$, where $\mathbf f_g=\left[ 0\ \ 0\ \ g\right]^{\text T} $, $g$ is the 
gravitational acceleration. The front two types of forces and moments will be described detailly in the following two subsections.

\subsubsection{Forces and Moments in the Quadcopter Component} 
In the quadcopter part, the thrust and torque produced by one rotor are given by
\begin{eqnarray}
	{{T}_{i}} =  {{K}_{f}}{{\varpi}_{i}}^{2}, {{M}_{i}} =  {{K}_{m}}{{\varpi}_{i}}^{2}=\frac{{K}_{m}}{{K}_{f}}{{T}_{i}}
\end{eqnarray}
where ${{K}_{f}>0}$ is the lift force coefficient, ${{K}_{m}>0}$ is the drag torque coefficient, and $\varpi_{i}$ is the angular 
rate of the \textit{i}th rotor, $i=1, 2, 3, 4$. In order to improve the controllability during performing yaw,
an installation angle $\eta$ is set as shown in Fig. \ref{Fig:8}(a). The left motors tilt to the positive left and right motors tilt to the 
positive right.

Because of the installation angle $\eta$, the forces and moments produced by rotors are expressed by the thrust on each propeller $T_{i}$ as
\begin{equation}
	\left[\begin{array}{c}
		f_{r_y}\\
		f_{r_z}\\
		{{m}_{r_x}}\\
		{{m}_{r_y}}\\
		{{m}_{r_z}}
	\end{array}\right]=\left[\begin{array}{cccc}
		\sin\eta & -\sin\eta & -\sin\eta & \sin\eta\\
		-\cos\eta & -\cos\eta & -\cos\eta & -\cos\eta\\
		-{{d}_{y}}\cos\eta & {{d}_{y}}\cos\eta & {{d}_{y}}\cos\eta & -{{d}_{y}}\cos\eta\\
		{{d}_{x}}\cos\eta & -{{d}_{x}}\cos\eta & {{d}_{x}}\cos\eta & -{{d}_{x}}\cos\eta\\
		K_{1} & K_{1} & -K_{1} & -K_{1}
	\end{array}\right]\left[\begin{array}{c}
		T_{1}\\
		T_{2}\\
		T_{3}\\
		T_{4}
	\end{array}\right],
\end{equation}
where $K_{1}={{K}_{m}}\left/{{K}_{f}}\right.+{{d}_{x}}\sin\eta$,
${{d}_{x}}$ and ${{d}_{y}}$ are the components of the distance from the center of the lifting-wing quadcopter to a propeller on the $o_{\rm b}x_{\rm b}y_{\rm b} $ plane, as shown in Fig. \ref{Fig:8}(c).

\begin{figure}
	\centering
	\includegraphics[scale=1.0]{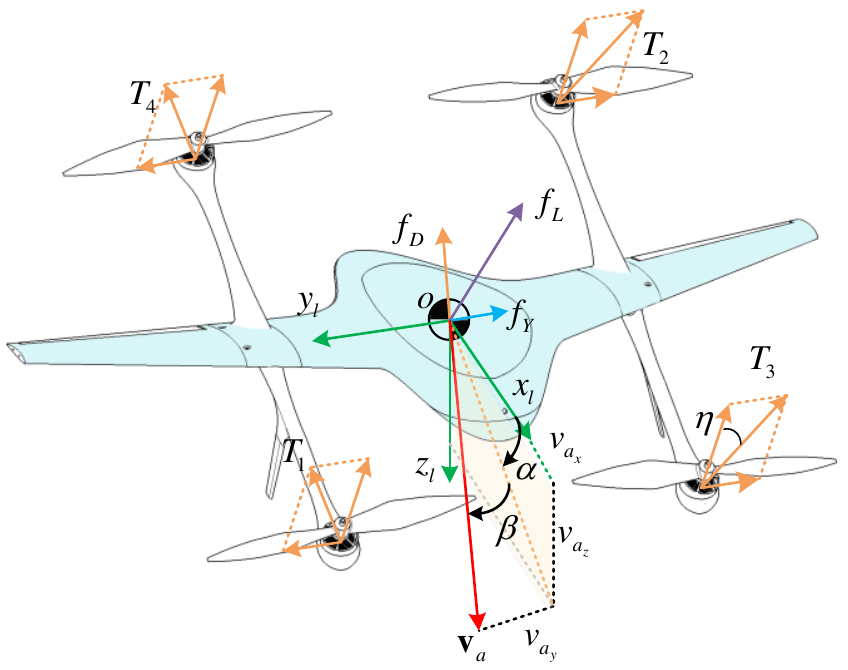}
	\caption{Forces act on a lifting-wing quadcopter.}
	\label{Fig:9}
\end{figure}

\subsubsection{Forces and Moments in the Lifting-wing Component}

The aerodynamic forces and moments acting on the lifting wing are mainly generated by the lifting wing itself and the ailerons at the trailing edge, 
as shown in Fig. \ref{Fig:9}.

Let $^{\rm{e}}\mathbf{v_{\rm{w}}}$ be the wind velocity in ${}^{\rm e}\mathcal{F} $. Then 
\begin{equation}
	^{\rm{b}}{{\mathbf{v}}_{\rm a}}={}^{\rm{b}}\mathbf{v}-({\bf {R}}_{{\rm {b}}}^{\rm e})^{\rm T}\cdot{}^{\rm{e}}{{\mathbf{v}}_{\rm{w}}}.
\end{equation}
Thus, the airspeed vector $^{\rm{l}}{{\mathbf{v}}_{\rm a}}=\left[{{v}_{{{\rm a}_x}}}\ \ {{v}_{{{\rm a}_y}}}\ \ 
{{v}_{{{\rm a}_z}}}\right]^{\text{T}}$
and airspeed $V_a$ are defined as 
\begin{equation}
	^{\rm{l}}{{\mathbf{v}}_{\rm a}}=({{\bf {R}}_{{\rm {l}}}^{\rm b}})^{\rm T}\cdot{}^{\rm{b}}{{\mathbf{v}}_{\rm a}}, 
\end{equation}
\begin{equation} 
	V_{\rm a}={\sqrt{{{v}^{2}_{{\rm a}_x}}+{{v}^{2}_{{\rm a}_y}}+{{v}^{2}_{{\rm a}_z}}}}.
\end{equation}
The aerodynamic angles $\alpha$ and $\beta$ are defined as
\begin{equation}
	\alpha={{\tan}^{-1}}(\frac{{{v}_{{\rm a}_z}}}{{{v}_{{\rm a}_x}}}),\ \beta={{\sin}^{-1}}(\frac{{{v}_{{\rm a}_y}}}{V_{\rm a}}).
\end{equation}

In the longitudinal plane, lift, drag, and pitching moment acting on the lifting-wing body are given by
\begin{equation}
	\begin{aligned}  {f_L}& = QS({{C}_{L}}+{{C}_{L{{\delta}_{{e}}}}}{{\delta}_{{e}}})\\
		{f_D}& = QS({{C}_{D}}+{{C}_{D{{\delta}_{{e}}}}}{{\delta}_{{e}}})\\
		m& = QSc({{C}_{m}}+{{C}_{m{{\delta}_{{e}}}}}{{\delta}_{{e}}}).
	\end{aligned}
	\label{Eq.LDm}
\end{equation}
The lateral force and the roll and yaw moments
acting on the lifting-wing body are given by
\begin{equation}
	\begin{aligned} 
		{{f}_{Y}}& =QS({{C}_{Y}}+{{C}_{Y{{\delta}_{a}}}}{{\delta}_{a}})\\
		l& =QSb({{C}_{l}}+{{C}_{l{{\delta}_{a}}}}{{\delta}_{a}})\\
		n& =QSb({{C}_{n}}+{{C}_{n{{\delta}_{a}}}}{{\delta}_{a}})
	\end{aligned}
\end{equation}
where $Q = \frac{1}{2}\rho V_a^2$; ${{C}_{L}}$, ${{C}_{D}}$, ${{C}_{m}}$,${{C}_{Y}}$, ${{C}_{l}}$
and ${{C}_{n}}$ are nondimensional aerodynamic coefficients, ${{C}_{L{{\delta}_{{e}}}}}$,
${{C}_{m{{\delta}_{{e}}}}}$, ${{C}_{D{{\delta}_{{e}}}}}$,${{C}_{Y{{\delta}_{a}}}}$,
${{C}_{n{{\delta}_{a}}}}$ and ${{C}_{l{{\delta}_{a}}}}$ are control
derivative; $S$ is the area of the lifting wing, $c$ is the mean chord
of the lifting wing, $b$ is the wingspan of the lifting-wing aircraft,
${{\delta}_{{e}}}$ and ${{\delta}_{ a}}$ are calculated
using the right and the left aileron (${{\delta}_{{a}r}}$ and
${{\delta}_{{a}l}}$, as shown in Fig. \ref{Fig:8}(d)) as 
\begin{equation}
	\left[\begin{matrix}{{\delta}_{{e}}}\\
		{{\delta}_{ a}}
	\end{matrix}\right]=\left[\begin{matrix}1 & 1\\
		-1 & 1
	\end{matrix}\right]\left[\begin{matrix}{{\delta}_{{a}r}}\\
		{{\delta}_{{a}l}}
	\end{matrix}\right].
\end{equation}

The external forces and moments are summarized as

\begin{align}
	{{}^{\rm{b}}\mathbf f}&=\left[\begin{matrix}0\\
		f_{r_y}\\
		f_{r_z}
	\end{matrix}\right]+QS{\mathbf{R}}_{\text{w}}^{\text{b}}\left[\begin{matrix}-({{C}_{D}}+{{C}_{D{{\delta}_{{e}}}}}{{\delta}_{{e}}})\\
		({{C}_{Y}}+{{C}_{Y{{\delta}_{ a}}}}{{\delta}_{ a}})\\
		-({{C}_{L}}+{{C}_{L{{\delta}_{{e}}}}}{{\delta}_{{e}}})
	\end{matrix}\right]+m\mathbf{R}_{\rm{e}}^{\rm{b}}\mathbf f_{g}
	\label{eq.f}\\
	{^{\rm{l}}\mathbf{m}}&={{\mathbf R}_{\text{b}}^{\text{l}}} \left[\begin{matrix}{{m}_{r_x}}\\
		{{m}_{r_y}}\\
		{{m}_{r_z}}
	\end{matrix}\right]+QS\left[\begin{matrix}b({{C}_{l}}+{{C}_{l{{\delta}_{ a}}}}{{\delta}_{ a}})\\
		c({{C}_{m}}+{{C}_{m{{\delta}_{{e}}}}}{{\delta}_{{e}}})\\
		b({{C}_{n}}+{{C}_{n{{\delta}_{ a}}}}{{\delta}_{ a}})
	\end{matrix}\right].
	\label{eq.m}
\end{align}
The structure parameters, lift force and drag torque coefficients of the lifting-wing quadcopter are given in Tab.\ref{Tab:coefficients}.
\renewcommand\arraystretch{1.2} 
\begin{table}
	\caption{Lifting-wing quadcopter structure parameters, lift force and drag torque coefficients }
	\begin{center}
		\begin{tabular}{c|c|c}
			
			\hline
			$m $    & Aircraft mass &1.92 kg \\	
			\hline	
			$\kappa$ &Installation angle of lifting wing &34  $ \rm deg$\\
			\hline
			$\eta $ & Installation angle of motor &10 deg \\
			\hline
			$d_x $  & The distance from $o_bx_b $ to a propeller&0.25 m\\
			\hline
			$d_y $  & The distance from $o_by_b $ to a propeller&0.2125 m\\
			\hline
			$ [J_{xx}\ \ J_{yy}\ \ J_{zz}]$  & Moment of inertia&$ [5.12\ \ 5.54\ \ 7.6]\times 10^{-2}$ $\rm kg\cdot m^2$\\
			\hline
			$b $    & Wingspan of the lifting-wing aircraft&0.94 $ \rm m$\\
			\hline
			$c $    & Mean chord of the lifting wing&0.17 $ \rm m$\\
			\hline
			$K_m $  & Drag moment coefficient& 5.875e-07 $ \rm kg\cdot m^2$\\
			\hline
			$K_f $  &Lift force coefficient &2.824e-05 $ \rm kg\cdot m^2$\\
			\hline
			
		\end{tabular}
	\end{center}
	\label{Tab:coefficients}
\end{table}
\renewcommand\arraystretch{0.8} 
\section{CONTROLLER DESIGN}
\begin{figure}
	\centering
	\includegraphics[scale=0.7]{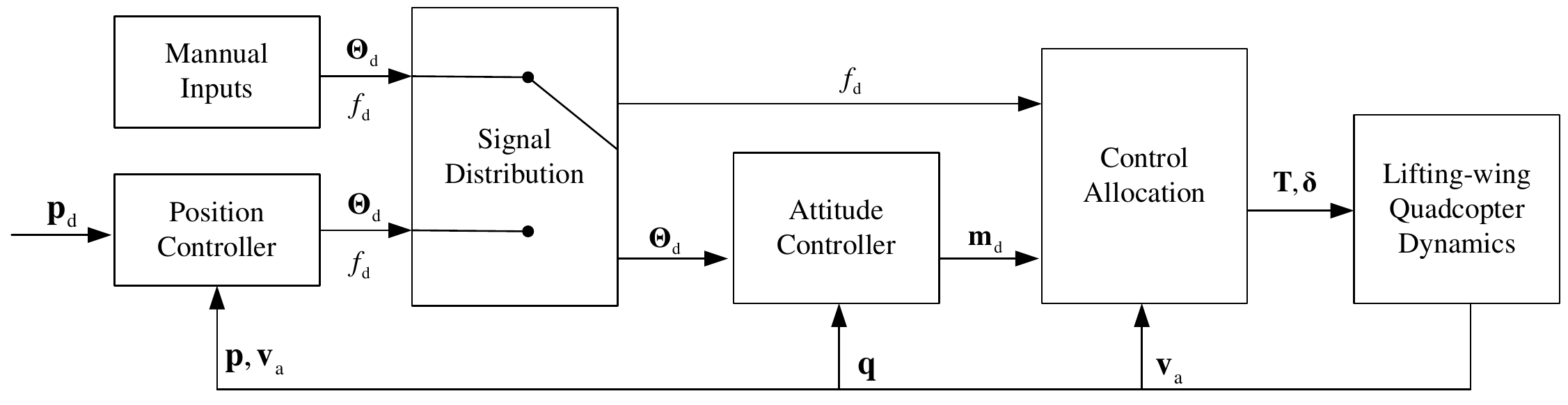}
	\caption{Control structure}
	\label{Fig: ControlStructure}
\end{figure}

The successive loop closure is a common control architecture for UAVs \cite{beard2012small}, which consists of an outer-loop controlling the  position and an inner-loop for attitude, as illustrated in Fig. \ref{Fig: ControlStructure}. The basic idea behind successive loop closure is to 
close several simple feedback loops in succession around the open-loop plant dynamics rather than designing a single control system. The position 
controller receives the desired position and then computes the desired acceleration. Then the desired acceleration is mapped to the collective thrust 
and attitude. The attitude command is sent to the inner-loop, while the thrust command skips directly to the control allocation. To facilitate the 
control experiment step by step, the attitude can also be commanded by the pilot. Furthermore, the attitude controller receives the desired attitude and 
generates the desired moment. Finally, the control allocation algorithm distributes the moment command from the inner-loop and the direct force command from the outer-loop to corresponding ailerons and rotors.

\subsection{Controller Design Model}

Since $\alpha,\beta$ are not easy to obtain, we consider that $\alpha\approx\kappa+\theta$
and $\beta\approx0$. The translational dynamic involves
five control variables, namely three-dimensional forces in frame ${}^{\rm b}\mathcal{F} $
and two Euler angles (pitch and roll angles). However, it is a bit too early
to determine the five control variables according to the three-dimensional
desired acceleration because it is hard to obtain the bounds of these
control variables. An improper choice may not be realized by ailerons.
To this end, we only choose $f_z$ (the main force component) in frame ${}^{\rm b}\mathcal{F} $ and two Euler angles (pitch and roll angles) to 
determine 
the desired acceleration uniquely, and the desired yaw angle is directly specified, leaving $f_y $ as a disturbance. This adopts the controlling idea 
of quadcopters\cite{quan2017introduction,nascimento2019position}, but the computation method is different due to the existence of the aerodynamic 
force. According to the idea above, we rewrite the system Eqs. \eqref{Eq:position kinematic and dynamic} and \eqref{Eq:attitude kinematic and dynamic} in the form as
\begin{equation} 
	\begin{aligned}
		{}^{\text{e}}{\dot{\mathbf p}} &= {}^{\text{e}}{\mathbf{v}} \\
		{}^{\text{e}}{\dot{\mathbf v}} &= {\mathbf{u}} + {\mathbf{g}} + {{\mathbf{d}}_1}  \\
		{\dot{\mathbf R}}_{\text{l}}^{\text{e}}&={\mathbf R}_{\text{l}}^{\text{e}}{\left[ {{}^{\text{l}}{\bm{\omega }}} \right]_ \times } \\
		{}^{\text{l}}{\bm{\dot {\omega} }} &= {{\mathbf{J}}^{ - 1}} \cdot {}^{\text{l}}{\mathbf{m}} + {{\mathbf{d}}_2}
	\end{aligned}.
	\label{simplemodel}
\end{equation}
Here 
\begin{equation*}
	{\mathbf{u}} = \frac{{{\mathbf{R}}_{\text{b}}^{\text{e}}}}{m}\left( {\left[ {\begin{array}{*{20}{c}}
				0 \\ 
				0 \\ 
				{ - {f_z}} 
		\end{array}} \right] + QS{\mathbf{R}}_{\text{w}}^{\text{b}}\left[ {\begin{array}{*{20}{c}}
				{ - {C_D}} \\ 
				0 \\ 
				{ - {C_L}} 
		\end{array}} \right]} \right),
\end{equation*}
and $\mathbf{d}_{1}$, $\mathbf{d}_{2}$ are disturbances, where
\begin{align}
	{{\mathbf{d}}_1}&=\frac{{{\mathbf{R}}_{\text{b}}^{\text{e}}}}{m} \left( 
	\left[\begin{array}{*{20}{c}}0\\
		{{f_y}} \\ 
		0 \end{array}\right]
	+QS{\mathbf{R}}_{\text{w}}^{\text{b}}\left[\begin{matrix}-{{C}_{D{{\delta}_{e}}}}{{\delta}_{e}}\\
		{({C_Y} + {C_{Y{\delta _{ a}}}}{\delta _{ a}})}\\
		-{{C}_{L{{\delta}_{{e}}}}}{{\delta}_{e}}
	\end{matrix}\right] \right) \nonumber
	\\
	{{\mathbf{d}}_2} &=  - {{\mathbf{J}}^{ - 1}} \cdot {}^{\text{l}}{\bm{\omega }} \times ({\mathbf{J}} \cdot {}^{\text{l}}{\bm{\omega }}).\nonumber
\end{align}

\subsection{Position Control} 
Given a twice differentiable trajectory$\ \mathbf{p}_{\text{d}}(t)$,
in order to satisfy $\underset{t\rightarrow\infty}{\lim}\left\Vert {}^{\text e}\mathbf{p}(t)-\mathbf{p}_{\text{d}}(t)\right\Vert =0,$
the desired $\mathbf{u}_{\text{d}}$ for Eq. (\ref{simplemodel}) can be designed as a PID controller in the form 
\begin{equation}
	\mathbf{u}_{\text{d}}=-\mathbf 
	g + {\ddot{\mathbf p}}_{\text{d}}-{{\mathbf K}_{{\mathbf{P}}_{\text{d}}}}\left({}^{\text e}\mathbf{v}-{\dot{\mathbf 
	p}}_{\text{d}}\right)-{{\mathbf{K}}_{{\mathbf{P }}_{\text{p}}}}\left({}^{\text e}\mathbf{p}-\mathbf{p}_{\text{d}}\right)-{{\mathbf{K}}_{{\mathbf{P }}_{\text{i}}}}\int\left({}^{\text 
	e}\mathbf{p}-\mathbf{p}_{\text{d}}\right)\text{d}s
	\label{ud}
\end{equation}
where ${{\mathbf{K}}_{{\mathbf{P }}_{\text{p}}}}$, ${{\mathbf{K}}_{{\mathbf{P }}_{\text{i}}}}$, ${{\mathbf K}_{{\mathbf{P}}_{\text{d}}}} \in 
{\mathbb{R}^{3 \times 3}}$ are diagonal matrices acting as control gains. The left work is to determine
desired thrust by rotors $f_{\text{d}}\in\Omega_{f}$ and $\theta_{\text{d}},\phi_{\text{d}}\in\Omega_{a}$
such that
\begin{equation}
	\left(f_{\text{d}},\theta_{\text{d}},\phi_{\text{d}}\right)=\underset{f_z \in\Omega_{f},\theta,\phi\in\Omega_{a}}{\arg\min}\left\Vert \mathbf{u}\left(f_z ,\theta,\phi\right)-\mathbf{u}_{\text{d}}\right\Vert
	\label{Eq30} 
\end{equation}
where $\Omega_{f}$ is a set to confine the force, and $\Omega_{a}$
is a set to confine the pitch and roll. In order to reduce drag, the vehicle's nose
should be consistent with the current direction of the vehicle velocity,
that is
\begin{equation}
	{\psi_{{\rm {d}}}}={\tan^{-1}}\left({\frac{{v_{{y_{{\rm {\rm{e}}}}}}}}{{v_{{x_{{\rm {\rm{e}}}}}}}}}\right).
\end{equation}

The attitude command can also be given by the pilot, in case the position controller fails with GPS denied, as shown in Fig. \ref{Fig: ControlStructure}. Finally, the desired attitude is given as ${{\mathbf{\Theta }}_{{\text{d}}}} = {\left[ {{\phi_{{\text{d}}}}}\ \ {{\theta _{{\text{d}}}} }\ \ \psi _{{\text{d}}} \right]^{\text{T}}}$.

\subsection{Attitude Control}
The attitude controller generates the desired moment from the output of the position controller or the attitude given by 
the pilot, as shown in Fig. \ref{Fig: ControlStructure}. As far as we know, studies about the hybrid UAV 
hardly consider the lateral control, such as turning right or left. As for the considered UAV, the control on yaw is quite different between the multicopter mode and the fixed-wing mode. To establish a unified control, the attitude control is performed on the lifting-wing frame ${}^{\rm l}\mathcal{F} $, so $^{\text{l}}{{\mathbf{\Theta }}_{{\text{d}}}} = {\left[ {{\phi_{\text {\text{d}}}}}\ \ {{\theta _{\text {\text{d}}}}+\kappa }\ \ \psi _{\text {\text{d}}} \right]^{\text{T}}}$.

\subsubsection{Basic Attitude Control}
The attitude error is presented in the form of quaternion based on which the corresponding controller is designed. This can guarantee a uniform and good convergence rate for all initial attitude errors \cite{lyu_hierarchical_2017}
\begin{equation}
	{\mathbf{q}_{\rm e}} = {\mathbf{q}_{\rm d}^{*}} \otimes {\mathbf{q_{\rm l}^{\rm e}}} = {\left[ q_{\text{e}_0}\ \ {q_{\text{e}_1}}\ \ 
	{q_{\text{e}_2}}\ \ {q_{\text{e}_3}} \right]^{\text{T}}},
\end{equation}
where ${\mathbf{q}}_{\text{d}}$ is transformed from $^{\text{l}}{{\mathbf{\Theta }}_{{\text{d}}}}$ with `ZXY' rotational sequence, $ (\cdot)^*$ is 
the conjugate of a quaternion, and $\otimes $ is the quaternion product. Then
${\mathbf{q}}_{\text{e}}$ is transformed into the axis-angle form ${\mathbf{q}}_{\text{e}}= \left[ \cos \frac{\vartheta}{2} \ \ {\bm \xi}^{\text T}_{\text e}\sin \frac{\vartheta}{2} \right]^{\text T}$ by 
\begin{equation}
	\begin{gathered}
		\vartheta  = {\text {wrap}}_\pi \left( {2{\text{acos}}\left( q_{\text{e}_0} \right)} \right) \hfill \\
		{\bm \xi}_{\text{e}}  = \left\{ {\begin{array}{{ll}}
				{{\left[ 0\ \ 0\ \ 0 \right]}^{\text{T}}},& \theta  = 0 \\ 
				{\text{sign}}(q_{\text{e}_0}) \frac{\vartheta }{{\sin {\vartheta  \mathord{\left/{\vphantom {\vartheta  2}} \right.
				\kern-\nulldelimiterspace} 2}}}{{\left[ {q_{\text{e}_1}\ \ q_{\text{e}_2}\ \ q_{\text{e}_3} }\right]}^{\text{T}}},&\theta  \ne 0 .
		\end{array}} \right. \hfill \\ 
	\end{gathered}
\end{equation}
The function ${\text {wrap}_\pi }(\vartheta)$  constrains the $ \vartheta$ in $\left[{-\pi\ \ \pi}\right]$ to ensure the shortest rotation path. To eliminate the attitude error, the attitude control is designed as 
\begin{equation}
	{}^{\text{l}}{{\bm{\omega }}_{\text{ac}}} = {\text {sat}}\left ( {{\mathbf{K }}_{{\mathbf{\Theta }}{\text{p}}}}{{\bm \xi} _{\text e}}, \bm \omega _{\text {min}},\bm \omega _{\text {max}} \right )
\end{equation}
where ${{\mathbf{K}}_{{\mathbf{\Theta }}{\text{p}}}} \in {\mathbb{R}^{3 \times 3}}$ is the diagonal matrix acting as the control gain, $\bm \omega _{\text {min}} $ and $\bm \omega _{\text {max}} \in {\mathbb{R}^{3}}$ are the minimum and maximum  angular control rates, the function $ \text {sat} \left (  \mathbf x ,\mathbf {x}_{\min},\mathbf {x}_{\max} \right)$ is defined as
\begin{equation}
	\text {sat} \left (  \mathbf x ,\mathbf {x}_{\min},\mathbf {x}_{\max} \right) \triangleq \left[ {\begin{array}{*{20}{c}}
			{{\text{sat}}\left( {{x_1},{x_{1,\min }},{x_{1,\max }}} \right)} \\ 
			\vdots  \\ 
			{{\text{sat}}\left( {{x_n},{x_{n,\min }},{x_{n,\max }}} \right)} 
	\end{array}} \right],
	{\text{sat}}\left( {{x_k},{x_{k,\min }},{x_{k,\max }}} \right) \triangleq \left\{ {\begin{array}{*{20}{ll}}
			{x_{k,\min }},&{x_k} < {x_{k,\min }} \\ 
			{x_{k,\max }},&{x_k} > {x_{k,\max }} \\ 
			{x_k},&{\text{else}} 
	\end{array}}. \right.
\end{equation}

\subsubsection{Lateral Compensation}
When the UAV is in high-speed flight, a roll command given to track a specified trajectory will cause a lateral skid. In order to reduce the sideslip angle when the UAV turns at a high speed, the coordinated turn should be considered. In ${}^{\rm 
l}\mathcal{F} $ frame, if it is assumed that there is no wind, the coordinated turn equation is expressed as \cite{beard2012small}
\begin{equation}
	{\dot \psi _{\text d}} = \frac{{g\tan \phi}}{{{V_a}}} \label{Turnsimple}.
\end{equation}
 It should be noted that the Euler angles are the attitude presentation between ${}^{\rm l}\mathcal{F} $. So the desired yaw rate generated by coordinated turn is 
\begin{equation}
 	^{\text{l}}{{\mathbf{\omega }}_{{\text{ct}}}} = {\dot \psi _{\text d}}\cos \theta \cos \phi  \label{Eq:deuler}.
\end{equation}

\subsubsection{Angular Rate Command Synthesis}
In the lifting-wing coordinated turn, $ V_a$ being zero makes no sense. In addition, considering that coordinated turn should not be used when airspeed is small, a weight coefficient related to airspeed is added. So the desired angular rates are rewritten as 
\begin{equation}
	^{\text{l}}{{\mathbf{\omega }}_{{\text{d}}}} = \left[ 
	{\mathbf{\omega }}_{{\text{d} _{\text{ac},x}}} \ \  {\mathbf{\omega }}_{{\text{d} _{\text{ac},y}}} \ \ \left({\mathbf{\omega }}_{{\text{d} _{\text{ac},z}}} + w{ \cdot ^{\text{l}}}{{\mathbf{\omega }}_{{\text{ct}}}} \right)
	\right]
\end{equation}
where $ $ $w = {\text{sat}}\left( {\frac{{{V_{a}} - {V_{\min }}}}{{{V_{\max }} - {V_{\min }}}},0,1} \right)$. When the airspeed is slower than $V_{\min} $, the desired yaw rate is completely decided by the basic attitude controller. In contrast, when ${ ^{\text{l}}}{{\mathbf{\omega }}_{{\text{d} _z}}}=0 $ and the airspeed reaches the specified value $V_{\max} $, the desired yaw rate is completely decided by the coordinated turn.

\subsubsection{Attitude Rate Control}
 To eliminate the attitude rate error, the controller is designed as
\begin{equation}
	\begin{gathered}
		{}^{\text{l}}{{\mathbf{m}}_{\text{d}}} = {\text{sat}}\left( {{\mathbf{J}}( - {{\mathbf{K}}_{{\bm{\omega 
		}}_{\text{p}}}}({}^{\text{l}}{\bm{\omega }} - {}^{\text{l}}{{\bm{\omega }}_{\text{d}}}) - {{\mathbf{m}}_{{\text{d,I}}}} - 
		{{\mathbf{K}}_{{\bm{\omega }}_{\text{d}}}}({}^{\text{l}}{\bm{\dot \omega }} - {}^{\text{l}}{{{\bm{\dot \omega }}}_{\text{d}}})), - 
		{{\mathbf{m}}_{{\text{d,max}}}},{{\mathbf{m}}_{{\text{d,max}}}}} \right) 
	\end{gathered}
\end{equation} 
where ${{\mathbf{m}}_{{\text{d,I}}}} = {\text{sat}}\left( {{{\mathbf{K}}_{{\bm{\omega }}_{\text{i}}}}\int {({}^{\text{l}}{\bm{\omega }} - 
		{}^{\text{l}}{{\bm{\omega }}_{\text{d}}}){\text{d}}s,}  - 
	{{\mathbf{m}}_{{\text{d,Imax}}}}, {{\mathbf{m}}_{{\text{d,Imax}}}}} 
\right) $, ${{\mathbf{K}}_{{\mathbf{\omega }}_{\text{p}}}}$, ${{\mathbf{K}}_{{\mathbf{\omega }}_{\text{i}}}}$, ${{\mathbf K}_{{\bm{\omega 
}}_{\text{d}}}} \in {\mathbb{R}^{3 \times 3}}$ are diagonal matrices acting as control gains, ${{\mathbf{m}}_{{\text{d,Imax}}}}$ is maximum amplitude of integral action, ${{\mathbf{m}}_{{\text{d,max}}}}$ is the maximum moment generated by actuators.

So far, we have obtained $f_{\text{d}}$ and ${}^{\text l}\mathbf{m}_{\text{d}}$, which will be further realized by $T_{1},\ \cdots,\ T_{4},\ 
\delta_{ {ar}},\ \delta_{{ {al}}}$.

\subsection{Control Allocation}
The lifting-wing quadcopter is an over-actuated aircraft, which provides six independent control inputs, namely $T_{1},\ \cdots,\ T_{4},\ \delta_{ {ar}},\ \delta_{{al}}$ to meet a specific thrust $f_{\text{d}} \in {\mathbb{R}}$ and moment demand ${}^{\rm l}\mathbf{m}_{\text{d}} \in {\mathbb{R}^{3}}$. A method of control allocation based 
on optimization is proposed. Recalling the control in translational dynamic, if we determined the three-dimensional force in ${}^{\rm b}\mathcal{F} $ 
and two Euler
angles (pitch and roll angles) by an optimization before, then the six control variables (plus desired yaw angle) will be determined by six actuators 
uniquely. If so, however, the optimization in the control of the translational dynamic is not related to energy-saving directly. This is why we only choose the 
$o_{\rm{b}}z_{\rm{b}}$ force (the main force component) and two Euler angles (pitch and roll angles) to determine the desired acceleration as in 
Eq.(\ref{Eq30}). 

First, Eqs. (\ref{eq.f}) and (\ref{eq.m}) are rearranged as
\begin{equation}
	\underbrace {\left[ {\begin{array}{*{20}{c}}
			{{f_z}}\\ 
			{{}^{\rm l}{m_x}} - {QSb{C_l}}\\ 
			{{}^{\rm l}{m_y}} - {QSc{C_m}}\\ 
			{{}^{\rm l}{m_z}} - {QSb{C_n}}
	\end{array}} \right]}_{\mathbf u_{\text v}} = \underbrace {\left[ {\begin{array}{*{20}{c}}
				{ - \cos \eta }&{ - \cos \eta }&{ - \cos \eta }&{ - \cos \eta }&0&0 \\ 
				{ - {K_2}}&{{K_3}}&{{K_2}}&{ - {K_3}}&{ - QSb{C_{l{\delta _a}}}}&{QSb{C_{l{\delta _{a}}}}} \\ 
				{{d_x}\cos \eta }&{ - {d_x}\cos \eta }&{{d_x}\cos \eta }&{ - {d_x}\cos \eta }&{QSc{C_{m{\delta _e}}}}&{QSc{C_{m{\delta _e}}}} \\ 
				{ - {K_5}}&{{K_4}}&{{K_5}}&{ - {K_4}}&{ - QSb{C_{n{\delta _a}}}}&{QSb{C_{n{\delta _a}}}} 
		\end{array}} \right]}_{\mathbf{B}}\underbrace {\left[ {\begin{array}{*{20}{c}}
				{{T_1}} \\ 
				{{T_2}} \\ 
				{{T_3}} \\ 
				{{T_4}} \\ 
				{{\delta _{{{ar}}}}} \\ 
				{{\delta _{{{al}}}}} 
		\end{array}} \right]}_{\bm{\delta }} 
	\label{controlAllocation}
\end{equation}
where ${K_2} = {d_y}\cos \eta \cos \kappa  + {K_1}\sin \kappa$, ${K_3} = {d_y}\cos \eta \cos \kappa  - {K_1}\sin \kappa$, ${K_4} = {d_y}\cos \eta 
\sin \kappa  + {K_1}\cos \kappa$, 
${K_5} = {d_y}\cos \eta \sin \kappa  - {K_1}\cos \kappa$. The value ${\bm{\delta }}$ is the control input of the actuator , ${{\mathbf{u}}_{\text v}} $ is the virtual control, $ \mathbf B$ is the control efficiency matrix. 

As shown in the Eq.\eqref{controlAllocation}, $ \text {rank}\left( \mathbf B\right)=4$, the dimension of $\bm{\delta } $ is 6, which is higher than that of ${\mathbf u_{\text v}}$, so Eq.\eqref{controlAllocation} has a minimum norm solution. To take the control priority of the ${\mathbf u_{\text v}}$ components, actuator priority of $\bm{\delta } $ and actuator saturation under consideration, the control allocation is formulated to be an optimization problem as
\begin{equation}
	\begin{gathered}
		\min \;{\left\| {{{\mathbf{W}}_{\mathbf{u}}}\left( {{\mathbf{B}\bm{\delta }} - {{\mathbf{u}}_{\text {v,d}}}} \right)} \right\|^2} + \gamma 
		{\left\| 
		{{{\mathbf{W}}_{\bm{\delta }}}\left( {{\bm{\delta }} - {{\bm{\delta }}_{\text p}}} \right)} \right\|^2} \hfill \\
		\text 
		{s.t.}\quad {\mathbf{\underset{\raise0.3em\hbox{$\smash{\scriptscriptstyle-}$}}{\bm\delta } }} \leqslant {\bm{\delta }} \leqslant 
		{\bm{\bar \delta }} \hfill 
		\label{controlAllocation1}
	\end{gathered} 
\end{equation}
where ${\mathbf{u}}_\text{v,d}=\left[ {{f_{\text d}}}\ \ {{}^{\rm l}{m_{\text d _x}} - QSb{C_l}}\ \ {{}^{\rm l}{m_{\text d_y}} - QSc{C_m}}\ \ {{}^{\rm l}{m_{\text d_z}} - QSb{C_n}} 
\right]^{\rm T}$ is the desired virtual control, ${{\bm{\delta }}_{\text p}}$ is the preferred control vector which will be specified later, $ {{\mathbf{W}}_{\mathbf{u}}} \in {\mathbb{R}^{6 \times 6}}$ is a positive definite weighting matrix that prioritizes the commands in case the desired virtual input ${\mathbf{u}}_\text{v,d}$ cannot be achieved, ${{\mathbf{W}}_{\bm{\delta }}} \in {\mathbb{R}^{6 \times 6}}$ is a positive definite weighting matrix that prioritizes the different actuators, ${\mathbf{\underset{\raise0.3em\hbox{$\smash{\scriptscriptstyle-}$}}{\bm\delta } }}{\text{ = }}\max 
({{\bm{\delta }}_{\min }},{{\bm{\delta }}_{\text l}} - \Delta {\bm{\delta }})$ , ${\bm{\bar \delta }}{\text{ = }}\min 
({{\bm{\delta }}_{\max }},{{\bm{\delta }}_{\text l}} + \Delta {\bm{\delta }})$ are lower and upper bounds at each sampling instant of actuators, ${{\bm{\delta }}_{\min }}$ and ${{\bm{\delta }}_{\max }}  \in {\mathbb{R}^{6}}$ are actuator position limits, $\Delta {\bm{\delta }} \in {\mathbb{R}^{6}}$ is the rate limit, and ${{\bm{\delta }}_{\text l}}$ is the last value of ${{\bm{\delta }}}$. 

There are two optimization objectives in Eq.\eqref{controlAllocation1}, namely ${\left\| {{{\mathbf{W}}_{\mathbf{u}}}\left( {{\mathbf{B}\bm{\delta }} - {{\mathbf{u}}_{\text {v,d}}}} \right)} \right\|^2} $ and ${\left\|{{{\mathbf{W}}_{\bm{\delta }}}\left( {{\bm{\delta }} - {{\bm{\delta }}_{\text p}}} \right)} \right\|^2}$. The first one  is the primary objective of minimizing the slack variables weighted by ${{\mathbf{W}}_{\mathbf{u}}}$, so the weighting factor $\gamma $ is often chosen to be a small value. In many cases, the preferred control vector is set to the last value of ${{\bm{\delta }}}$, ${{\bm{\delta }}_{\text p }}={{\bm{\delta }}_{\text l}}$. But, in order to save energy, we prefer to use the aerodynamic force because the change of rotor force implies the motor's rotational speed change, which is power-hungry. 
According to the consideration above, we can give more weight to $\delta _{ar}$ and $\delta _{al} $ and set first four elements of ${{\bm{\delta }}_{\text 
		p}}$ to $ \frac {T_1 + T_2 + T_3 + T_4}{4} $, and last two elements to that of ${{\bm{\delta }}_{\text l}}$.

The hardware resources of the flight control are very limited, so it is very important to ensure the real-time operation of the proposed algorithm. Several different algorithms, like redistributed pseudoinverse, interior-point methods, and active set methods have been proposed to solve the constrained quadratic programming problem. Among them, active set methods\cite{2013Control} perform well in the considered control allocation problems, because they have the advantage that their initialization can take advantage of the solution from the previous sample (known as the warm start), which is often a good guess for the optimal solution at the current sample. This can reduce the number of iterations needed to find the optimal solution in many cases. The study \cite{2003Efficient} shows that the computational complexity of the active set methods is similar to the redistributed pseudoinverse method  and the fixed-point algorithm, but the active set methods produce solutions with better accuracy.

\section{Simulation Experiments} \label{section:SIMULATION SETUP}

In order to verify the three main advantages of the designed aircraft: control performance, energy saving and reliable transition flight, 
several experiments are conducted. The verification of these performances is mainly determined by two factors, namely, with and without ailerons, and with and without coordinated turn. 

Therefore, experiments are primarily composed of two parts. One is the comparison of control performance, energy saving and transition flight with and without aileron. And the other is the comparison of control performance with and without coordinated turn, but both with aileron. In addition, the transition flight of three different VTOL vehicles, as shown in Fig. \ref{Fig:2}, is analyzed in the HIL simulation environment.

\subsection{Simulation Platform}

The HIL simulation is carried out in the RflySim platform \cite{wang2021rflysim,dai2021rflysim,rflysim.com}, which provides CopterSim simulator and supports Pixhawk/PX4 autopilot.When perform the HIL simulation, CopterSim sends sensor data such as accelerometer, barometer, magnetometer, which is generated by a mathematical model, to the Pixhawk system via a USB serial port. The Pixhawk/PX4 autopilot will receive the sensors data for state estimation by the EKF2 filter and send the estimated states to the controller through the internal uORB message bus as the feedback. Then the controller sends the control signal of each motor as output back to CopterSim. Thereby a close-loop is established in the HIL simulation. Compared with the numerical simulation, the control algorithm is deployed and run in a real embedded system as the real flight does. After HIL simulation, the controller will be directly deployed to a real vehicle and further verification experiments will be performed.
\begin{figure}[thpb]
	\centering
	\includegraphics[scale=0.4]{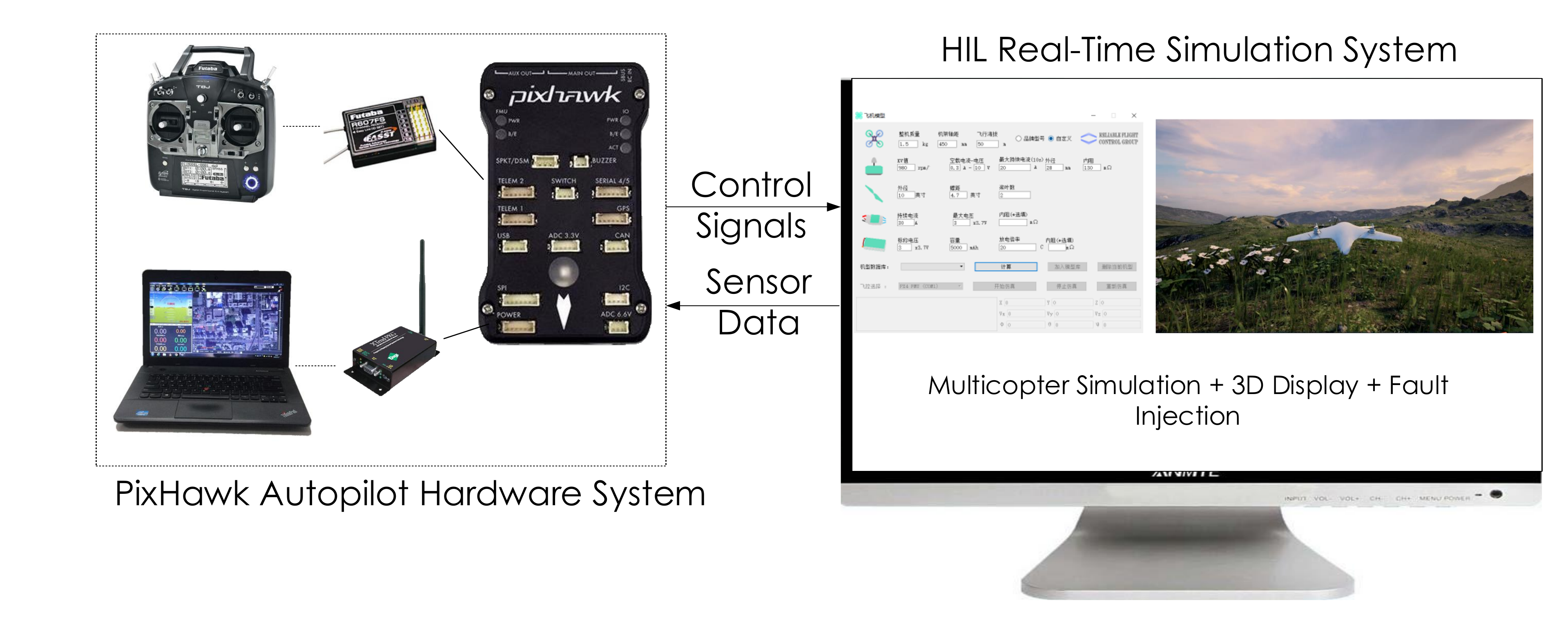}
	\caption{HIL simulation of RflyLW2.}
	\label{Fig:20}
\end{figure}

The lifting-wing quadcopter model is built in MATLAB/Simulink according to Eqs. \eqref{Eq:position kinematic and dynamic},\eqref{Eq:attitude 
kinematic and dynamic} \eqref{eq.f}, \eqref{eq.m}.
Besides, the dynamics of the motor and servo are modeled as first-order transfers as follows:
\begin{equation}
	{\varpi}=\frac{1}{T_{\text{m}}s+1}(C_\text{m}\sigma _i +{\varpi}_\text{b}),{\delta _\text a}=\frac{1}{T_{{\text a}}s+1}(C_\text{a}\sigma _j +{\delta}_\text{b}),i=1,2,3,4, j=5,6
	\label{Eq:lowpass}
\end{equation}
where $\sigma _i \in [0, 1]$ is $i \text {th}$ throttle command, $T_{\text{m}} $ and $T_{\text{a}} $ are the time constant parameter of motor and servo response, $C_\text{m}$, $C_\text{a}$, $\delta _\text{b}$ and ${\varpi}_\text{b}$ are constant parameters. The sensors used in HIL simulation include IMU, magnetometer, barometer, GPS and airspeed meter, are modeled reference to \cite{quan2017introduction}. 

\begin{figure}[thpb]
	\centering
	\includegraphics[scale=0.8]{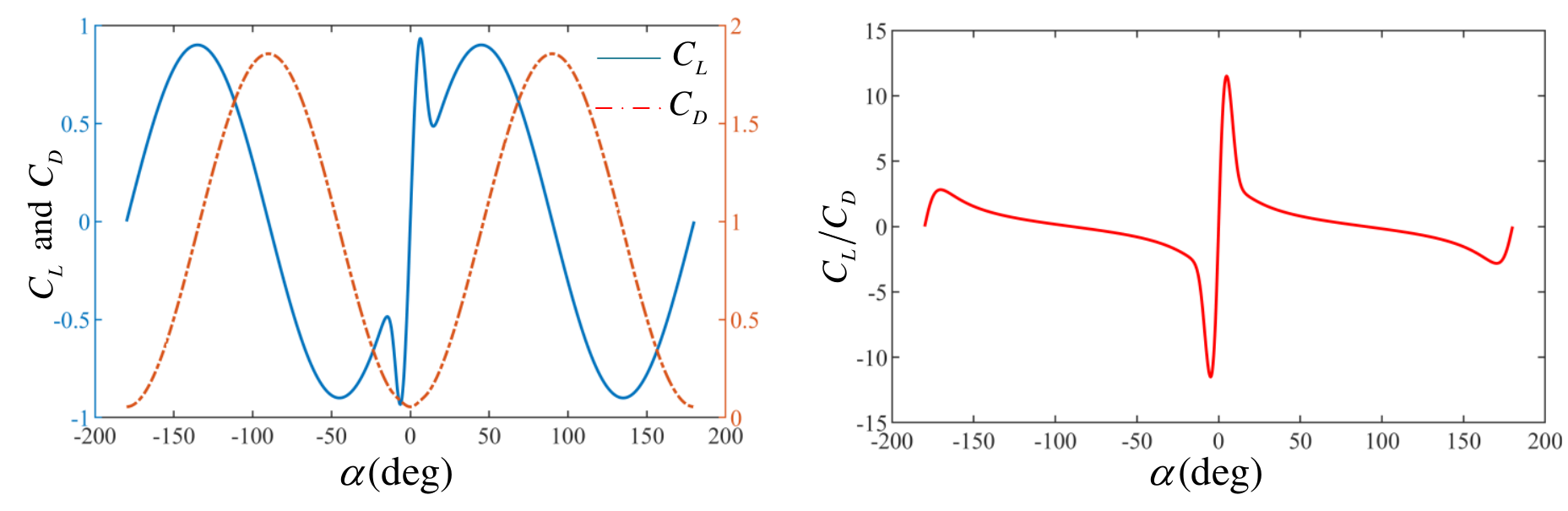}
	\caption{Aerodynamic parameters obtained by CFD}
	\label{Fig:Aerodynamic parameters}
\end{figure}

Due to the vehicle’s VTOL capabilities, the aerodynamic parameters must not only be modeled up to the stall AoA, but also in the post-stall region, in order to cover the entire flight envelope. The full angle aerodynamic parameters are obtained by combining the small angle aerodynamic parameters obtained by CFD simulation and the empirical aerodynamic model \cite{pucci2011nonlinear}. The aerodynamic characteristics at low and large AoA are approximated as
\begin{equation*}
\left\{ \begin{gathered}
	{C_{{L_S}}}(\alpha ) = \frac{{0.5c_2^2}}{{({c_2} - {c_3}){{\cos }^2}(\alpha ) + {c_3}}}\sin (2\alpha ) \hfill \\
	{C_{{D_S}}}\left( \alpha  \right) = {c_0} + \frac{{{c_2}{c_3}}}{{\left( {{c_2} - {c_3}} \right){{\cos }^2}\left( \alpha  \right) + {c_3}}}{\sin ^2}\left( \alpha  \right) \hfill \\ 
\end{gathered}  \right. ,
\left\{ \begin{gathered}
	{C_{{L_L}}}\left( \alpha  \right) = {c_1}\sin \left( {2\alpha } \right) \hfill \\
	{C_{{D_L}}}\left( \alpha  \right) = {c_0} + 2{c_1}{\sin ^2}\left( \alpha  \right) \hfill \\ 
\end{gathered}  \right. ,
\end{equation*}
and a pseudo-sigmoid function 
\begin{equation*}
	\sigma \left( {{\alpha _0},k,\alpha} \right) = \frac{{1 + \tanh \left( {k\alpha _0^2 - k{\alpha ^2}} \right)}}{{1 + {\text{tanh}}\left( {k\alpha _0^2} \right)}},\alpha  \in \left[ { - \pi ,\pi } \right)
\end{equation*}
is used to blend low and large AoA regions together
\begin{equation}
	\left\{ \begin{gathered}
		{C_L}\left( \alpha  \right) = {C_{{L_S}}}\left( \alpha  \right)\sigma \left( {{\alpha _0},{k_L},\alpha } \right) + {C_{{L_L}}}\left( \alpha  \right)\left[ {1 - \sigma \left( {{\alpha _0},{k_L},\alpha } \right)} \right] \hfill \\
		{C_D}\left( \alpha  \right) = {C_{{D_S}}}\left( \alpha  \right)\sigma \left( {{\alpha _0},{k_D},\alpha } \right) + {C_{{D_L}}}\left( \alpha  \right)\left[ {1 - \sigma \left( {{\alpha _0},{k_D},\alpha } \right)} \right] \hfill \\ 
	\end{gathered}  \right. .
	\label{Eq:Aerodynamic parameters}
\end{equation}
The low and large AOA parameters $c_0, c_1, c_2, c_3 $, and blending parameters $ \alpha _0, k_L, k_D$ are turned according to CFD results, and their values are set to $c_0=0.055, c_1=0.9, c_2=13.0, c_3=3.3,\alpha _0=3 \text {deg}, k_L=38, k_D=48 $. Corresponding Aerodynamic curves are shown in Fig. \ref{Fig:Aerodynamic parameters}.

\begin{figure}[t]
	\centering
	\includegraphics[scale=0.93]{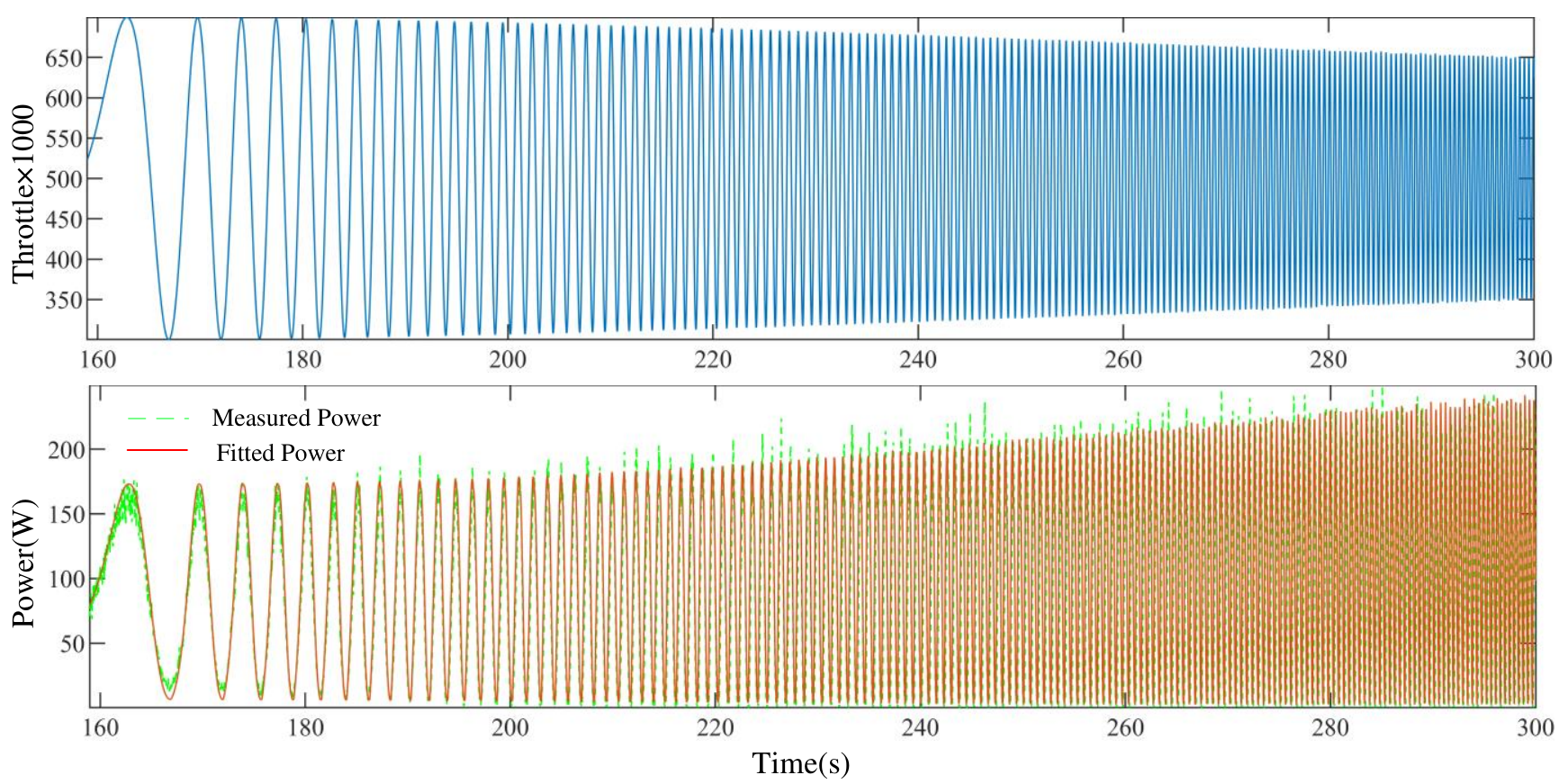}
	\caption{Throttle command and identification result of the motor}
	\label{Fig:motorIdenfication}
\end{figure}

\subsection{Simulation Experiments}
\subsubsection{Verifying the Effectiveness of Aileron}
\begin{figure}[thp]
	\centering
	\includegraphics[scale=0.7]{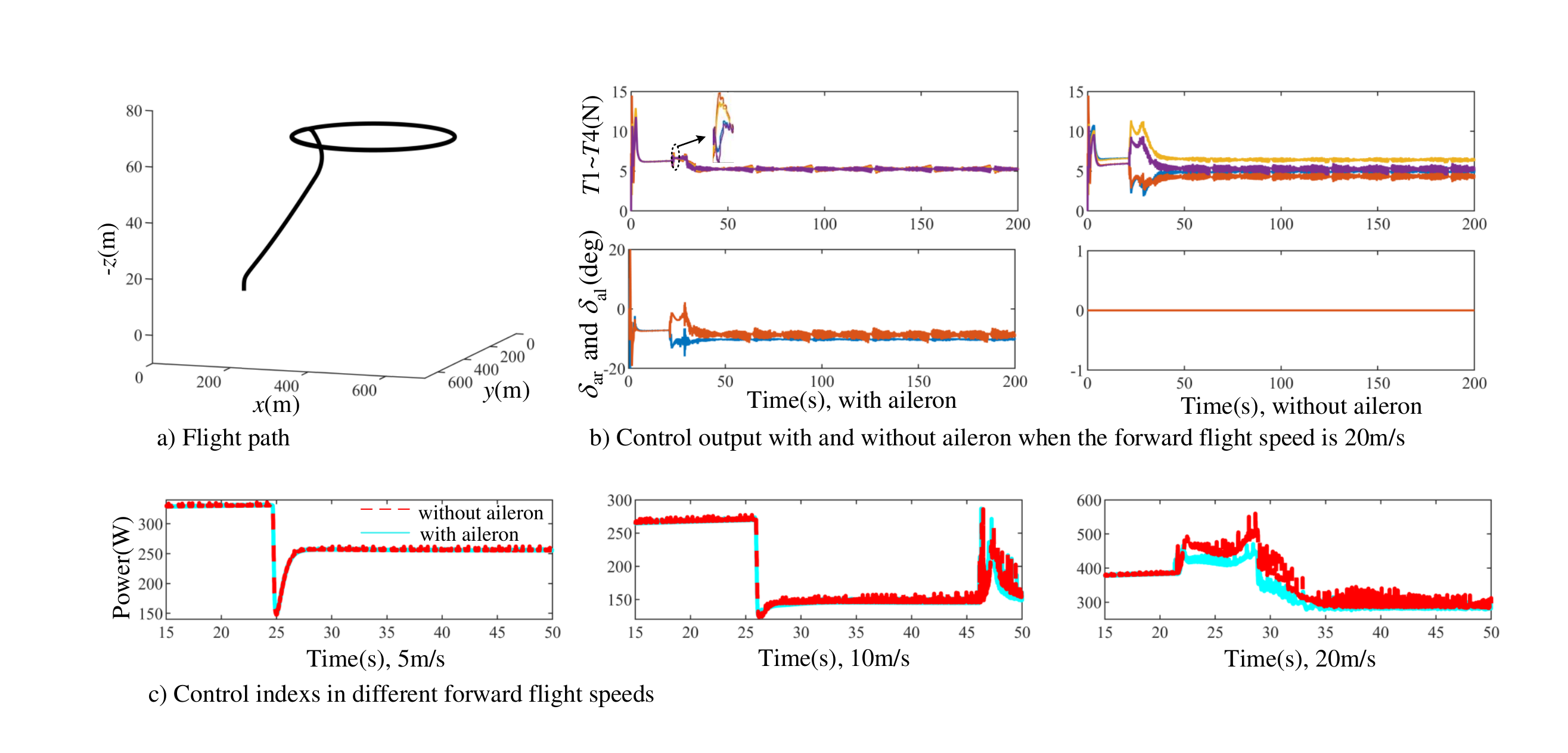}
	\caption{The attitude control performance and mixer output with and without aileron}
	\label{Fig:ControlPerformance}
\end{figure}

In order to show the effectiveness of ailerons, two groups of comparative experiments are carried out. The attitude control and 
position control are the same, but the control allocation of the first group uses the aileron, while the second group does not. The second group only depends on the quadcopter control allocation. In this experiment, the aircraft tracks the specified trajectory, as shown in Fig. \ref{Fig:ControlPerformance}(a), which is a straight line plus a circle with a radius of 200m.

As shown in Fig. \ref{Fig:ControlPerformance}(b), after adding the aileron, the control amplitude of the four motors is almost the same trend, especially when the aircraft transits between straight and circular trajectories, indicating that the attitude control is mainly realized by the aileron, and  motors are more like a thruster on a fixed-wing in this case. However, when the attitude changes sharply, as shown in Fig. \ref{Fig:ControlPerformance}(b) during $t=22\sim23$s, the motor and aileron will participate in the control at the same time to improve the control performance. When the state of the UAV is in the slow adjustment, the aileron undertakes the control as much as possible to save energy.

In order to quantify the influence of ailerons, the relationship between throttle command and energy consumption of the motor and actuator is identified. A sinusoidal signal with the constant amplitude and linear increasing frequency is designed, and the energy consumption test experiments are carried out on the servo and motor at the same time. Measurement and identification results of the motor are shown in the Fig.\ref{Fig:motorIdenfication}, in which the amplitude of the throttle decreases as the frequency increasing, because a low-pass filter in Eq.\eqref{Eq:lowpass} is applied to it. We find that the motor power has a quadratic function relationship with a fixed throttle, and the power is different when the motor accelerates and decelerates. When the change frequency of the throttle command becomes faster, the power is increased. Therefore, the following formula is established to fit the relationship between throttle command and motor power
\begin{equation}
	P_T=\sum\limits_{i=1}^{4} {p_1{\sigma _i}^2+p_2{\sigma _i} + p_3\dot \sigma _{up, i}^{p4} + p_5\dot \sigma _{down, i}^{p6} +p_7}
\end{equation}
With the given lifting-wing quadcopter platform, $ p_1 = 563.7, p_2=-147.4, p_3 = 15, p_4 = 1.05, p_5 = 4, p_6=1, p_7=0.05538$. With respect to the servo, its power is negligible, because the power of servo is only 0.2W, far less than the motor's, even if the 100g weight is suspended on the two ailerons.

The trajectory tracking experiment is performed three times, with the forward flight speed of 5m/s, 10m/s and 20m/s, respectively. The powers in three flight speeds are shown in  Fig. \ref{Fig:ControlPerformance}(c). It can be observed that when the speed is 5m/s, the power is the same, because the aileron is not used at slow flight speed. When the speed increases to 10m/s, the power with the aileron is slightly smaller. Further when the speed increases to 20m/s, the power with the aileron is greatly smaller because the aerodynamic force is stronger at high speed.

\subsubsection{Verifying the Effectiveness of Coordinated Turn}

\begin{figure}[t]
	\centering
	\includegraphics[scale=1.0]{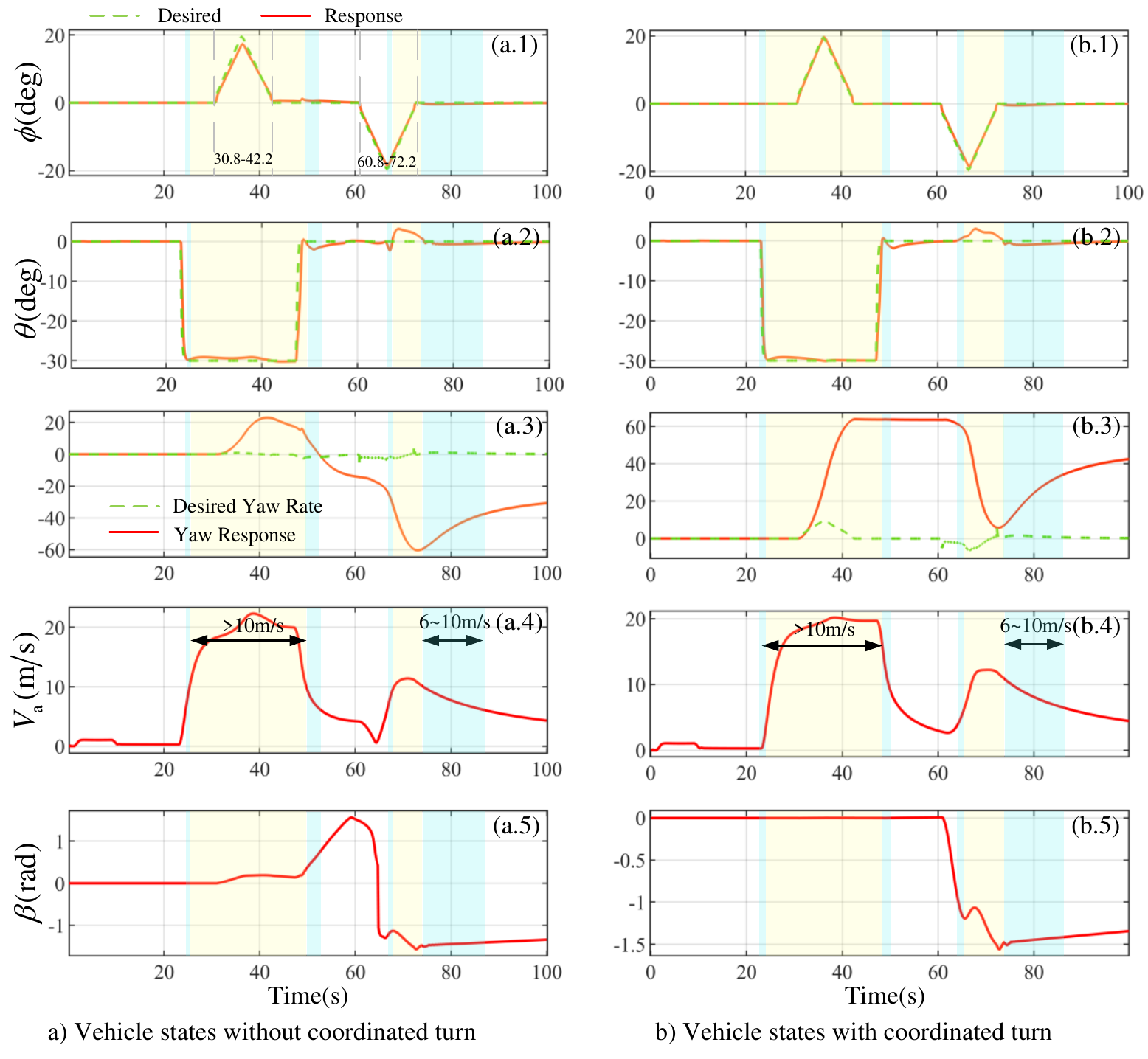}
	\caption{The control response with and without coordinated turn}
	\label{Fig:ControlPerformanceCT}
\end{figure}

As for the lateral control, experiments are carried out in hover and high-speed forward flight, and the results are shown in Fig. 
\ref{Fig:ControlPerformanceCT}. In the high-speed forward flight, as shown in Fig. \ref{Fig:ControlPerformanceCT} yellow region, when a turn command is given for the first time during $30.8\sim42.2$s, the lifting-wing quadcopter with coordinated turn has no sideslip angle and has a better lateral tracking performance, as shown in Fig.\ref{Fig:ControlPerformanceCT}(a.1 vs. b.1, and a.5 vs. b.5). In this state, the control of the lifting-wing quadcopter is similar to the fixed-wing, and the desired yaw rate is generated by Eq.\eqref{Eq:deuler} as the feedforward. When the turn command is given for the second time during $60.8\sim72.2$s, the lifting-wing quadcopter is in low-speed flight, and the control of the lifting-wing quadcopter is similar to the quadcopter, so a big sideslip angle appears both in experiments with and without coordinated turn.

\subsubsection{Transition Flight}

The transition flight is often defined as the aircraft changing from hover to level flight. To quantify the phase, the transition process is defined as the time it takes for the aircraft to start a transitional flight to the airspeed greater than 18m/s. The quality of the transition is mainly reflected in the \textit{transition time} and \textit{control accuracy of altitude}. In the simulation, the aircraft will take off to a certain altitude, and then a step signal of $ -30^\circ$ is given to the pitch channel. This is because the wind CFD test results show that the maximum energy efficiency is obtained when $ \alpha=4^\circ$. As a comparison, the same experiment is performed on a tail-sitter quadcopter. The model of the tail-sitter quadcopter is built on the lifting-wing quadcopter with the installation angle of the lifting-wing from $ 34^\circ$ to $ 90^\circ$, as shown in Fig. \ref{Fig:2}(a). 

\begin{figure}
	\centering
	\includegraphics[scale=0.75]{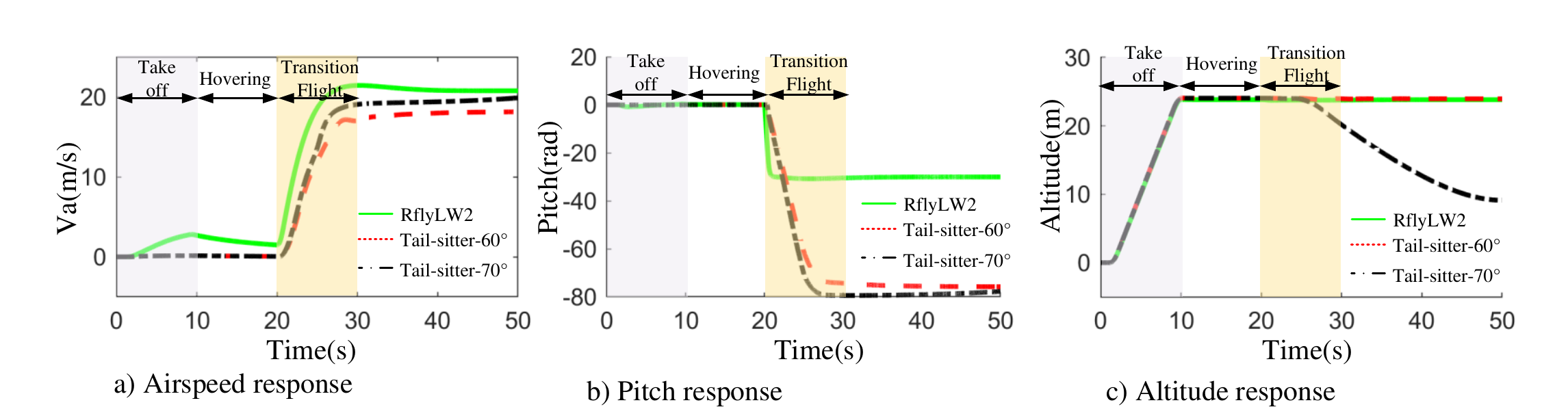}
	\caption{Transition flight of lifting-wing quadcopter and tail-sitter UAV.}
	\label{Fig:13}
\end{figure}

Fig.\ref{Fig:13} shows the response curves of the pitch angle, altitude and airspeed of the lifting-wing quadcopter and tail-sitter UAV in the transition phrase in the HIL simulation. It can be observed that during the transition phase of the lifting-wing quadcopter, the time for the pitch angle to reach the desired value is 1.1s, as shown in Fig.\ref{Fig:13}(b), and the time it takes for the airspeed to reach the set 18m/s is 4.7s as shown in Fig.\ref{Fig:13}(a). So the transition time of the lifting-wing quadcopter is 4.7s. Furthermore, the altitude decreases as soon as the transition starts, with a maximum altitude error of 0.09 m at $t$ = 21.68 s. As for tail-sitter UAV, the transition time is 7.1s, when the desired pitch angle is $ -70^\circ$. But after the transition, the altitude drops sharply, as shown in Fig.\ref{Fig:13}(c). When the desired pitch angle is $ -60^\circ$, the altitude is stable, but the transition time is 20.48s much longer than that of the RflyLW2. Thus, the transition phase of the lifting-wing quadcopter is significantly better than the tail-sitter UAV, and does not require an additional controller.
\section{Conclusion}

The modeling, controller design, and HIL simulation verification of the lifting-wing quadcopter—a novel type of hybrid aircraft—are presented in this paper. The modeling portion takes into account the forces and moments produced by the lifting wing and rotors. A unified controller for the entire flight phase is created based on the existing model. Depending on the velocity command, the controller can regard hovering and forward flight equally and enable a seamless transition between the two modes. The experimental results show that the proposed aircraft outperforms the tail-sitter UAV in terms of tracking performance during the transition phase. In addition, the controller combines the characteristics of the quadcopter and fixed-wing control law, allowing it to retain yaw during the hover phase and decrease sideslip angle during the cruise phase. What is more, a control allocation based on optimization is utilized to realize cooperative control for energy saving, by taking rotor thrust and aerodynamic force under consideration simultaneously. Through identification, we find that compared with the motor, the aileron can reduce the energy consumption when implementing high-frequency control inputs.
\bibliography{mylib}
\end{sloppypar}
\end{document}